\documentclass[final,3p,times,twocolumn]{elsarticle}
\usepackage{graphicx}%
\usepackage{multirow}%
\usepackage{amsmath,amssymb,amsfonts}%
\usepackage{amsthm}%
\usepackage{mathrsfs}%
\usepackage[title]{appendix}%
\usepackage{xcolor}%
\usepackage{textcomp}%
\usepackage{manyfoot}%
\usepackage{booktabs}%
\usepackage{algorithm}%
\usepackage{algorithmicx}
\usepackage{algpseudocode}%
\usepackage{listings}%
\usepackage{times}
\usepackage{epsfig}
\usepackage{bigstrut,multirow}
\usepackage{indentfirst}
\usepackage{url}
\usepackage{array}
\usepackage{float}
\usepackage{natbib}
\usepackage{balance}
\usepackage{etoolbox}
\usepackage{colortbl}
\usepackage{subcaption} 
\usepackage{caption}
\usepackage{wrapfig}
\usepackage[T1]{fontenc}
\makeatletter
\def\algocf@caption@algocf{\textbf{end while}}
\makeatother
\algnewcommand\algorithmicinput{\textbf{Input:}}
\algnewcommand\Input{\item[\algorithmicinput]}
\algnewcommand\algorithmicoutput{\textbf{Output:}}
\algnewcommand\Output{\item[\algorithmicoutput]}
\algnewcommand\algorithmicinitialization{\textbf{Initialization:}}
\algnewcommand\Initialization{\item[\algorithmicinitialization]}

\journal{Pattern Recognition}

\begin{document}

\begin{frontmatter}



\title{Distilled Transformers with Locally Enhanced Global Representations for Face Forgery Detection}

\author[label1]{Yaning Zhang}
\affiliation[label1]{organization={Computer Vision Institute, College of Computer Science and Software Engineering},
             addressline={ Shenzhen University},
             city={Shenzhen},
             postcode={518060},
             state={Guangdong},
             country={China}}

\affiliation[label2]{organization={National Engineering Laboratory for Big Data System Computing Technology},
             addressline={Shenzhen University},
             city={Shenzhen},
             postcode={518060},
             state={Guangdong},
             country={China}}
\affiliation[label3]{organization={Shenzhen Institute of Artificial Intelligence and Robotics for Society},
	city={Shenzhen},
	postcode={518129},
	state={Guangdong},
	country={China}}
\affiliation[label4]{organization={Guangdong Key Laboratory of Intelligent Information Processing},
	addressline={Shenzhen University},
	city={Shenzhen},
	postcode={518060},
	state={Guangdong},
	country={China}}
	\affiliation[label5]{organization={School of Computing and Information Technology},
		addressline={Great Bay University},
		city={Dongguan},
		postcode={523000},
		state={Guangdong},
		country={China}}
\author[label1,label2]{Qiufu Li}
\author[label5]{Zitong Yu}
\author[label1,label2,label3,label4]{\corref{cor1}Linlin Shen}
\cortext[cor1]{Corresponding author.}
\ead{llshen@szu.edu.cn}
\begin{abstract}
Face forgery detection (FFD) is devoted to detecting the authenticity of face images. Although current CNN-based works achieve outstanding performance in FFD, they are susceptible to capturing local forgery patterns generated by various manipulation methods. Though transformer-based detectors exhibit improvements in modeling global dependencies, they are not good at exploring local forgery artifacts. Hybrid transformer-based networks are designed to capture local and global manipulated traces, but they tend to suffer from the attention collapse issue as the transformer block goes deeper. Besides, soft labels are rarely available. In this paper, we propose a distilled transformer network (DTN) to capture both rich local and global forgery traces and learn general and common representations for different forgery faces. Specifically, we design a mixture of expert (MoE) module to mine various robust forgery embeddings. Moreover, a locally-enhanced vision transformer (LEVT) module is proposed to learn locally-enhanced global representations. We design a lightweight multi-attention scaling (MAS) module to avoid attention collapse, which can be plugged and played in any transformer-based models with only a slight increase in computational costs. In addition, we propose a deepfake self-distillation (DSD) scheme to provide the model with abundant soft label information. Extensive experiments show that the proposed method surpasses the state of the arts on five deepfake datasets.
\end{abstract}



\end{frontmatter}


\section{Introduction}
\label{}

Deepfakes are edited multimedia generated using deep learning-based approaches such as autoencoders (AE), generative adversarial networks (GAN). Although deepfake can aid the automation of game design, film production, and face generation, it may slander personal reputations, raise political crises, and threaten social stability if utilized for evil purposes. Especially with the recent development of generation technology, synthetic facial videos are becoming more and more realistic. The above challenges have driven progress in face forgery detection (FFD) using deep neural networks. 

\begin{figure}[t!]%
	\includegraphics[width=\linewidth]{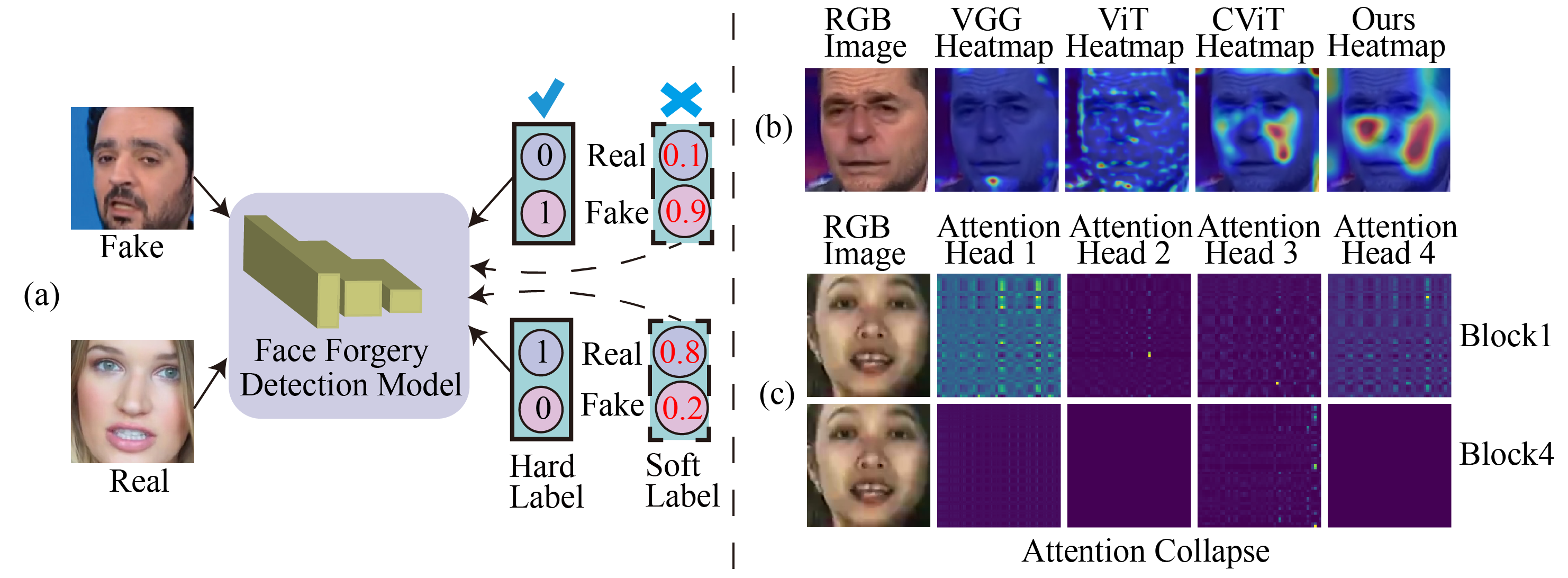}
	\centering
	\captionsetup{labelsep=none}
	\caption{\ (a) Schematic illustration of face image manipulation labels (real or fake). (b) The heatmap visualization of various deepfake detectors. (c) The visualization of attention maps across various heads in different transformer blocks.}\label{fig23}
\end{figure}

  Most traditional face forgery detectors \citep{wodajo2021deepfake,yu} are supervised only through a hard manipulation label (real or fake) to mine forgery traces. As Figure~\ref{fig23} (a) shows, \textbf{ A soft label such as the possibility of real or fake is rarely available in the FFD task.} FFD models may reach a bottleneck to further enhance detection performance due to the limited information entropy.
Current FFD detectors mainly involve three categories, i.e., CNN-based models, transformer-based methods, and hybrid transformer-based networks. Most CNN-based detectors \citep{yu,CHEN2023109179} may not capture global forgery patterns. As shown in Figure~\ref{fig23}, Visual Geometry Group (VGG) \citep{vgg} solely studies local features, which tends to fail to mine comprehensive manipulated patterns. Transformer-based models \citep{Alexey2021An,shao2022detecting} are introduced to explore long-range manipulation traces, but they are not good at mining local forgery artifacts. As Fig.~\ref{fig23} (b) shows, vision transformer (ViT) \citep{Alexey2021An} merely focuses on global counterfeiting traces at a coarse-grained level, which may struggle to capture subtle forgery artifacts due to a lack of prior knowledge related to the image such as vision-specific inductive biases. Hybrid transformer-based models \citep{wodajo2021deepfake,coccomini2022combining} such as convolutional vision transformer (CViT) \citep{wodajo2021deepfake} are designed to combine both local and global features, to explore comprehensive and fine forgery regions. \textbf{ We observe that current hybrid transformer-based networks applied for FFD tasks are prone to suffer from attention collapse when the transformer block goes deeper.} As Figure~\ref{fig23} (c) displays, the attention map of each head in a deeper block focused on fewer regions, and attention maps across heads tend to resemble each other with the number of blocks increasing, which is insufficient to capture more diverse comprehensive forgery patterns.

\begin{figure}[t!]%
	
	\includegraphics[width=\linewidth]{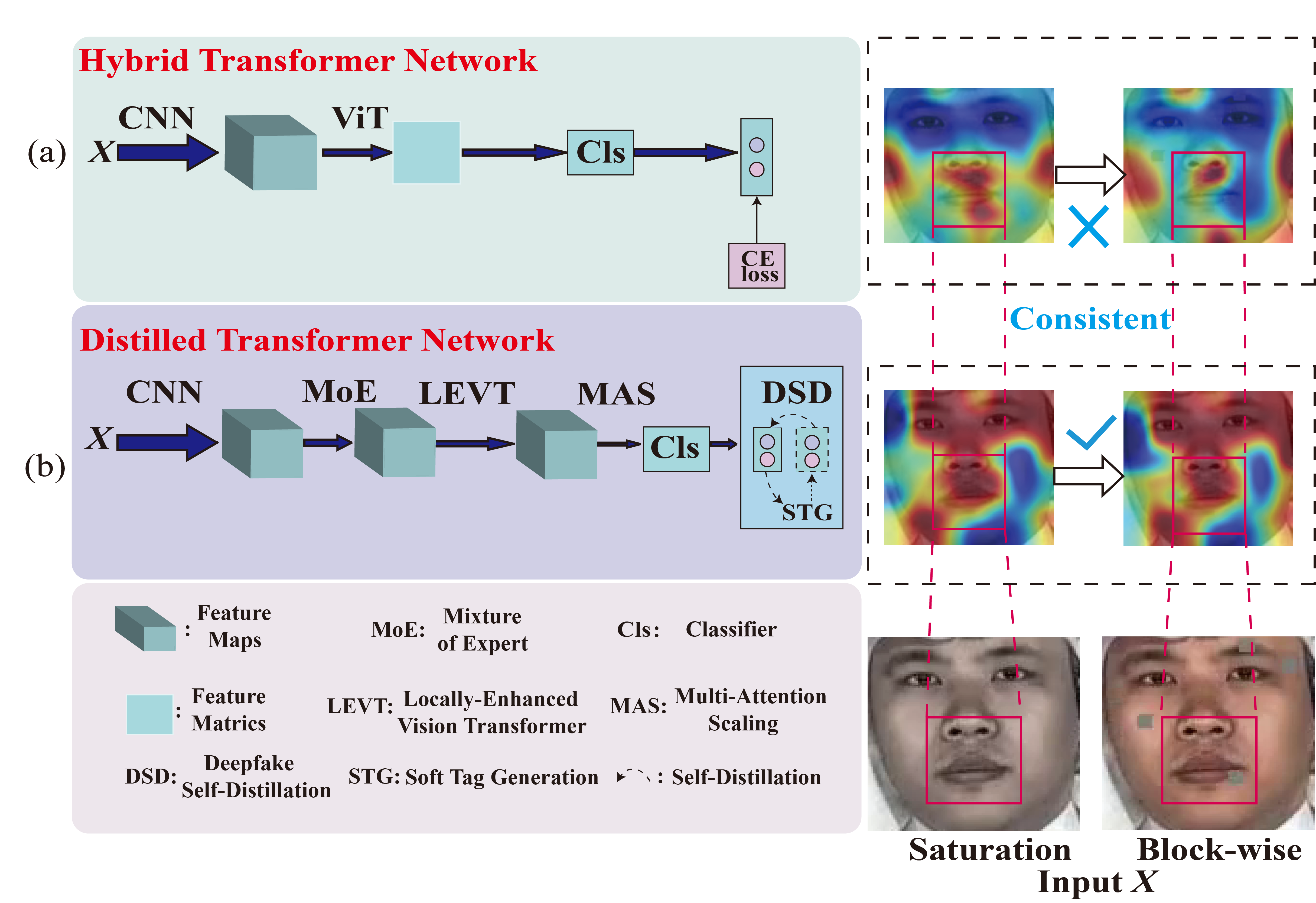}
	\centering
	\captionsetup{labelsep=none}
	\caption{\ (a) Traditional hybrid transformer, i.e. Convolutional vision transformer (CViT). (b) The proposed distilled transformer network (DTN). The heatmap of CViT and DTN for detecting faces in shifted domain attacks (saturation and block-wise are selected from seven types of image corruptions in \citep{Jiang2020DeeperForensics}). CViT struggles to mine the consistent forgery artifacts while DTN is capable of capturing the consistent and comprehensive forgery patterns.}\label{fig2}
\end{figure}

Based on aforementioned discussions, we mainly tackle the following issues. \textbf{Problem (1)}: How to provide more soft label information such as the possibility of real or fake for supervision to facilitate the advancement of FFD. \textbf{Problem (2)}: How to address the issue of attention collapse, thus learning more general long-range manipulation traces based on prevailing hybrid transformer-based models. To overcome these limitations, we propose a distilled transformer network (DTN) framework for FFD. As shown in Figure~\ref{fig2}, DTN tends to explore more inherent counterfeiting traces in different contexts. DTN differs from traditional hybrid transformers \citep{wodajo2021deepfake,coccomini2022combining} in the following aspects: Aiming at the problem (1), inspired by knowledge distillation (KD), we design a soft tag generation (STG) module to produce soft labels, which are generated by teacher models without the requirement of manual annotation. We further propose the deepfake self-distillation (DSD) scheme to build a chain of teacher-student discriminators, where the student and teacher have the same structure, and the student will serve as the teacher in the subsequent generation until no improvements are observed. Furthermore, We propose a mixture of expert (MoE) module to capture diverse dependencies among patches, and learn various robust forgery embeddings using multiple experts. We also devise an innovative locally-enhanced vision transformer (LEVT) module to study locally-enhanced global representations, thus exploring finer-grained and general manipulation artifacts globally. To solve the limitation of the attention collapse, we design a multi-attention scaling (MAS) method to adaptively select attention maps, to enhance the diversity of attention maps, thus exploring rich general forgery traces. To sum up, the contributions of our work are as follows:

$\bullet$ We design a MoE module to extract various robust forgery representations via the flexible integration of multiple expert decisions, and a LEVT module to capture locally-enhanced global dependencies, which contribute to exploring general and comprehensive forgery traces.

$\bullet$ We devise a plug-and-play MAS module to increase the diversity of attention maps via adaptively choosing attention maps, and thus avoid attention collapse, which can be integrated into any transformer-based models with only a slight growth in computational costs.

$\bullet$ We propose a novel deepfake self-distillation (DSD) scheme that yields soft labels via the STG module and combines metric learning with self-distillation, which can be applied to any detectors for general forgery pattern extraction. 

$\bullet$ Extensive experimental results compared with the state-of-the-art models show the superior effectiveness and generalizability of our proposed model on five deepfake datasets.

\section{Related Works}\label{sec2}

\subsection{Face Forgery Detection}\label{subsec21}

Some deep learning networks have been designed to alleviate the security risk caused by deepfakes. To learn the discriminative information of facial regions, Luo et al. \citep{luo2021generalizing} leveraged a two-stream (TwoStream) network to combine the high-frequency feature and RGB content for face manipulation detection. Recently, Cao et al. \citep{cao2022end} designed a reconstruction-classification learning (RECCE) network, which utilizes a CNN-based encoder and decoder to reconstruct real images for more general face representations, and constructs multi-scale bipartite graphs based on encoder output and decoder embeddings for deepfake detection. Pu et al. \citep{ pu2022learning} devised a dual-level collaborative model, to mine frame-level and video-level forgery traces simultaneously using a joint loss function. Chen et al. \citep{CHEN2023109179} treated FFD as a multi-task paradigm, to identify both intrinsic and extrinsic forgery artifacts simultaneously. Shiohara et al. \citep{Shiohara_2022_CVPR} blended the source and target images with the masks to create self-blended images (SBIs), and trained CNN-based models with SBIs to explore generic representations. The above methods hardly take the global knowledge between local patches into account. To overcome this limitation, a convolutional vision transformer (CViT) framework \citep{wodajo2021deepfake} is leveraged to combine CNN with a vision transformer (ViT) \citep{Alexey2021An} to detect authenticity. Coccomini et al. \citep{coccomini2022combining} integrated CNN-based backbones with various vision transformers, yielding the efficient vision transformer (EfficientViT) and cross-attention efficient vision transformer (CrossEfficientViT) for FFD. Dong et al.\citep{IDDDM} designed an ID-unaware deepfake detection model (IDDDM) to decrease the effect of the ID representation. Huang et al. \citep{IID} proposed an implicit identity-driven (IID) network to discern forgery faces with unknown manipulations. Shao et al. \citep{shao2022detecting} devised the SeqFakeFormer to combine the CNN with a transformer for detecting sequential deepfake manipulation. By contrast, our proposed approach can generate soft labels automatically for FFD, and is capable of suppressing the attention collapse and learning diverse robust features as well as convolution-enhanced global embeddings. Furthermore, deepfake self-distillation learning is exploited to further improve generalization performance.
\subsection{Mixture of Expert}\label{subsec28}
The mixture of experts (MoE) approach intends to divide complex problems into simpler sub-problems, which are handled by specialized expert models.
Wang et al. \citep{Divide} split the seen data into semantically independent subsets, and leveraged corresponding experts to study the semantics for zero-shot dialogue state tracking. He et al. \citep{Merging} designed a computation-efficient approach, which merges experts into one, to reduce the computational cost. Liu et al. \citep{diversifying} devised the OMoE optimizer and an alternative training scheme, to improve the diversity among experts in MoE for language models.
\subsection{Transformers in Computer Vision}\label{subsec22}

Recently, transformers \citep{Alexey2021An, WisdoMIM} have been used to solve computer vision issues such as image classification and object detection. For example, the large vison-language transformer such as WisdoMIM \citep{WisdoMIM} is proposed to conduct multi-modality learning and reasoning.

ViT \citep{Alexey2021An} regards an image patch as a word and first realizes the application of the transformer in the computer vision domain. To explore multi-modality learning and reasoning, WisdoMIM \citep{WisdoMIM} utilized contextual world knowledge from large vision-language models to improve multi-head self-attention (MHSA), and presented a training-free contextual fusion mechanism to minimize context noises. Wang et al. \citep{MitigatingHI} contrasted distributions of standard and instruction perturbations, to amplify alignment uncertainty and minimize hallucinations during large vision-language model inference. CanLK \citep{CanLK} is a large-scale multimodal alignment probing benchmark for current vision-language pre-trained models.

Different from other visual tasks such as image classification and objective detection, deepfake detection is more complicated and can be regarded as a fine-grained pattern recognition problem, since the artifact traces generated by advanced forgery methods are relatively subtle. Different manipulations produce various distributions, so learning general embeddings becomes increasingly challenging. In our work, we try to explore hybrid transformers in FFD tasks. 

\subsection{Knowledge Distillation}\label{subsec28}
Knowledge distillation (KD) is originally designed to transfer knowledge from a teacher model to a student network. The primary concept is that the student can imitate the output of the teacher. Recently, Zhong et al. \citep{RevisitingKD} found that various tokens have different teaching methods, and devised an adaptive
	teaching scheme to yield more diverse and flexible teaching modes. Rao et al. \citep{Parameter} introduced an adapter module specifically designed for the teacher model, where only the adapter is updated to generate soft labels with proper smoothness. Ding et al. \citep{Ding2020UnderstandingAI} introduced a Kullback-Leibler divergence scheme to compare the model lexical choices with those in the raw data, to preserve the valuable information in low-frequency words. Previous effort \citep{yang2019training} has demonstrated that students can still profit even if the teacher and student have the same architectures. Inspired by these works, we designed the soft tag generation module, where detailed labels are yielded by the teacher model automatically without the requirement of manual annotation. We further proposed the DSD scheme, where the student of one generation is employed as the teacher of the following generation. The procedure of DSD learning is repeated until no further performance improvements. 
\begin{figure*}[t!]
	\centering
	\includegraphics[width=\linewidth]{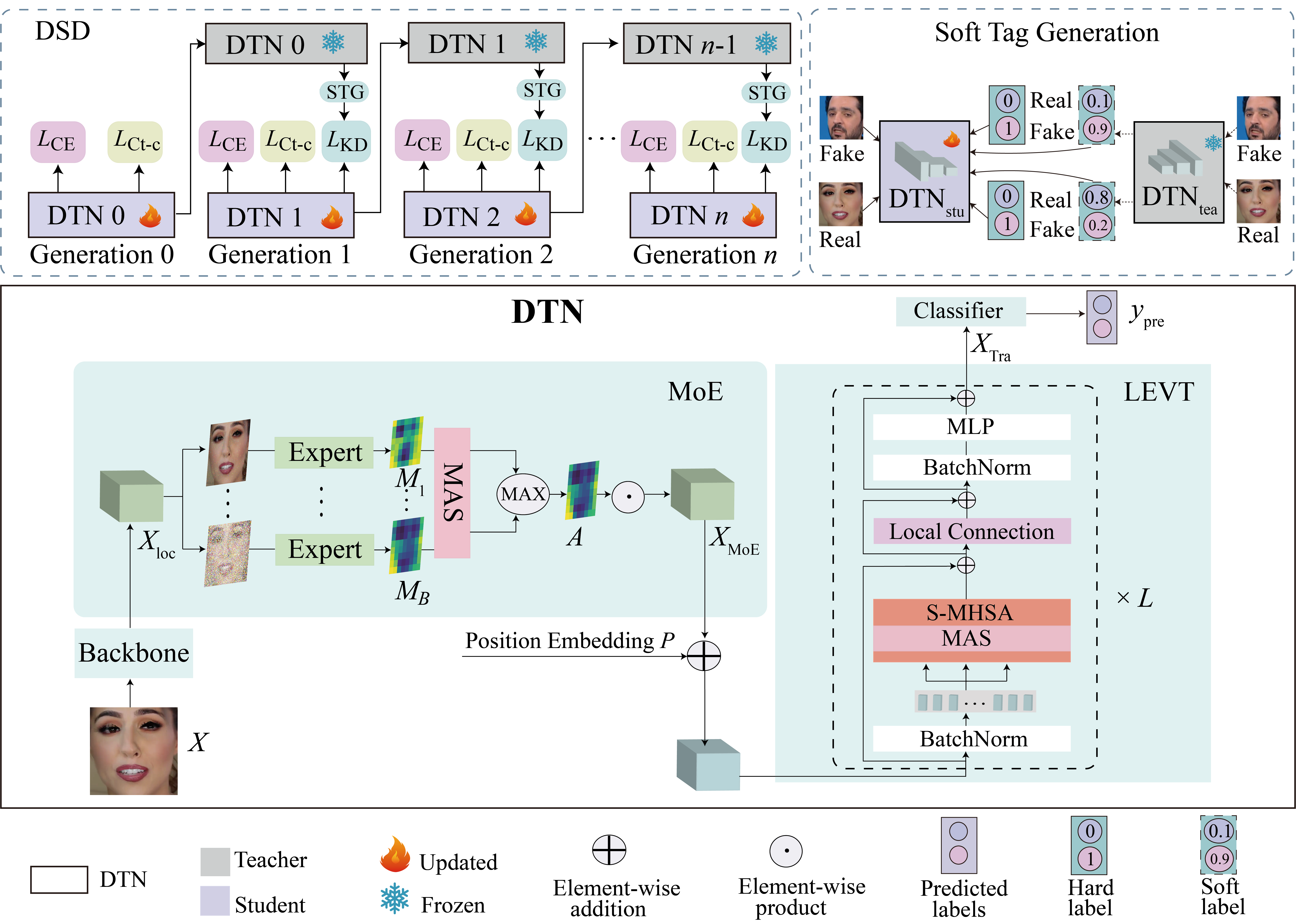} 
	\captionsetup{labelsep=none}
	\caption{\ An overview of the proposed DTN framework. We utilize the DSD scheme to capture general forgery traces, where we first pre-train a DTN model with labeled training images as the teacher model, to guide the student model learning, and one generation of the student serves as the teacher in the next one until no improvements are observed. The DTN model encodes high-level semantic embeddings from input facial images through a backbone VGG, which are then fed into the MoE module to analyze various robust embeddings. We then transfer them to the LEVT module to model the locally enhanced global relations among image patches, where MAS further mines rich facial forgery patterns via flexibly choosing attention maps. Finally, the classifier yields predictions.}\label{fig3}
\end{figure*}

\section{Method Overview}\label{sec_method_overview}
In this work, we propose the distilled transformed-based network (DTN) for FFD, 
which adopts deepfake self-distillation (DSD) strategy to explore the general forgery artifacts in the facial images. 

\subsection{Initial Training}\label{sec_init_training}
As Figure~\ref{fig3} shows, in the DSD strategy, the DTN is first trained on a deepfake detection dataset $D$ from scratch, 
supervised by the cross-entropy (CE) loss $L_{\text{CE}}$ and low-dimensional contrastive-center (Ct-c) loss $L_{\text{Ct-c}}$.
In this initial training, for each facial image $X\in D$ with one-hot label $y\in\big\{[0,1]^T,[1,0]^T\big\}$ annotated by manually, 
DTN generates the predicted label $y_{\text{pre}}\in \mathbb R^{2\times1}$, i.e.,
\begin{align}
	y_{\text{pre}} & = \text{DTN}(X),
\end{align}
where $y_{\text{pre}}$ denotes the probabilities, i.e., logits, of the input image $X$ belonging to real and fake.
Then, the losses are
\begin{align}
	L_{\text{CE}}(X) &= -y^T\log(y_{\text{pre}}),\\
	L_{\text{Ct-c}}(X) &= \frac{1}{2}\frac{\|y_{\text{pre}} - c_{y}\|^2}{\|y_{\text{pre}}-c_{1-y}\|+\varepsilon},
\end{align}
where $c_y$ represents the learnable center logits of the category to which image $X$ belongs,
$c_{1-y}$ denotes that of the other class, and $\varepsilon$ is a constant to avoid zero denominators. 

To minimize the distance from the embeddings of real and fake data to respective class center points, while pushing natural faces away from manipulated ones, we utilize the low-dimensional Ct-c loss. We compute the $L_{\text{Ct-c}}$ using the low-dimensional logits, i.e., the predicted probabilities generated by the model DTN, 
instead of high-level semantic features applied in the vanilla Ct-c loss \citep{contrastive}, which helps the model DTN to learn more generalized representations.

\begin{algorithm}
	\caption{Procedure of the Self-Distillation}\label{algo1}
	\begin{algorithmic}[1]
		\Input {The pretrained teacher model $\text{DTN}_{\text{tea}}(X;t)$; Student model with the same structure as the teacher $\text{DTN}_{\text{stu}}(X;s)$; Minimum loss value $\mathcal{L}_\textrm{min}$; The number of times the loss does not decrease $z$; The threshold t of the $z$; Distillation metric $\mathcal{L}_\text{tea}$; Training start flag $u$.}
		
		\Output {The state-of-the-art student model $\text{DTN}_{\text{stu}}^{\text{sota}}$.}
		\Initialization {$\mathcal{L}_\textrm{min}$$\leftarrow$${10}^6$; $z$$\leftarrow$0; $u$$\leftarrow$1.}

		\While{$\mathcal{L}_{\text{stu}}$$<$$\mathcal{L}_{\text{tea}}$ \textbf{or} $u$}
		\State $\text{DTN}_{\text{stu}}(X;s)$$\leftarrow$$\text{DTN}_{\text{tea}}(X;t)$	
		\State $\text{DTN}_{\text{tea}}^{\text{copy}}(X;t)$	$\leftarrow$$\text{DTN}_{\text{tea}}(X;t)$	
		\State The trained teacher model $\text{DTN}_{\text{tea}}(X;t)$ remains frozen	 
		\State Calculate the DSD loss $\mathcal{L}_{\text{stu}}$ for the student model by Eq.~(\ref{eq7})	 
		\State $\mathcal{L}_{\text{tea}}$$\leftarrow$$\mathcal{L}_{\text{stu}}$
		\State $u$$\leftarrow$0
		\While{$z$$<$t}	    
		\If{$\mathcal{L}_{\text{stu}}$$<$$\mathcal{L}_\textrm{min}$}
		\State $\mathcal{L}_\textrm{min}$$\leftarrow$$\mathcal{L}_{\text{stu}}$
		\State Save the student model $\text{DTN}_{\text{stu}}(X;s_z)$
		\Else 
		\State 	$z$$\leftarrow$$z+1$ 	           
		\EndIf
		\EndWhile
		\State Obtain the student model $\text{DTN}_{\text{stu}}(X;s_z)$      
		\State $\text{DTN}_{\text{tea}}(X;t)$$\leftarrow$$\text{DTN}_{\text{stu}}(X;s_z)$ \        
		\State $\mathcal{L}_{\text{stu}}$$\leftarrow$$\mathcal{L}_\textrm{min}$\       
		\State $\mathcal{L}_\textrm{min}$$\leftarrow$${10}^6$\       
		\State $z$$\leftarrow$0
		\EndWhile
		\State $\text{DTN}_{\text{stu}}^{\text{sota}}$$\leftarrow$$\text{DTN}_{\text{tea}}^{\text{copy}}(X;t)$
		\State\Return {$\text{DTN}_{\text{stu}}^{\text{sota}}$}
	\end{algorithmic}
\end{algorithm}

\subsection{Self-distillation}
After the initial training, we perform $n$ self-distillations on the model DTN.
As Figure~\ref{fig3} shows, in each self-distillation, we take the previously trained DTN as the teacher model, $\text{DTN}_{\text{tea}}$,
which is also taken as the student model, $\text{DTN}_{\text{stu}}$. We then yield the soft label automatically using the SFG module. In the self-distillation training, the parameters of the student could be updated via back propagation, while those of the teacher are frozen. We utilize the DSD loss to train the student model. The whole self-distillation procedure of the DTN model is outlined in Algorithm~\ref{algo1}.

{\bfseries\setlength{\parindent}{1.3em} Soft tag generation.} Different from most detectors that are supervised through coarse-grained hard labels to study forgery patterns, we design the STG module to create soft labels automatically, thus providing the detector with more information. As shown in Figure~\ref{fig3}, given a teacher model $\text{DTN}_{\text{tea}}$, and a learnable student model $\text{DTN}_{\text{stu}}$, we fed face images $X$ to the frozen $\text{DTN}_{\text{tea}}$, to obtain soft labels $y_{\text{pre}}^{\text{tea}}$. Meanwhile, $X$ is fed into $\text{DTN}_{\text{stu}}$ to get predicted tags $y_{\text{pre}}^{\text{stu}}$. 
\begin{align}
	y_{\text{pre}}^{\text{tea}} &= \text{DTN}_{\text{tea}}(X)\\
	y_{\text{pre}}^{\text{stu}} &= \text{DTN}_{\text{stu}}(X)
\end{align}

{\bfseries\setlength\parindent{1.3em} DSD loss.} To make $y_{\text{pre}}^{\text{stu}}$ and $y_{\text{pre}}^{\text{tea}}$ close in the label space, we devise the KD loss as follows:
\begin{align}
	L_{\text{KD}}(X) &= \delta(y_{\text{pre}}^{\text{tea}})^T\log\frac{\delta(y_{\text{pre}}^{\text{tea}})}{\delta(y_{\text{pre}}^{\text{stu}})},
\end{align}
where $\delta$ is the softmax function with temperature.

However, the soft label $y_{\text{pre}}^{\text{tea}}$ may contain wrong information, i.e., a fake image is marked as a real one, which tends to mislead model learning. We further employ manually annotated hard labels $y$ to guide the student network, to learn correct knowledge. To narrow the gap between $y$ and $	y_{\text{pre}}^{\text{stu}}$ in the label space, we utilize the CE loss. However, the learned features are not general enough since CE only focuses on the decision boundary between classes, rather than intra-class compression and inter-class separability. Consequently, we introduce Ct-c loss for metric learning. Finally, we design the DSD loss, to minimize the combination of CE loss for $y$, KD loss for $y_{\text{pre}}^{\text{stu}}$, and Ct-c loss for intra-class compactness and inter-class separability. Therefore, the total deepfake self-distillation (DSD) loss is
\begin{align}
	L_{\text{DSD}} = \alpha_1 L_{\text{CE}} + \alpha_2 L_{\text{Ct-c}} + \alpha_3 L_{\text{KD}},\label{eq7}
\end{align} 
where $\alpha_1$, $\alpha_2$, and $\alpha_3$ are three balance factors, and $L_{\text{CE}}(X)$ and $L_{\text{Ct-c}}(X)$ are computed using the student predicted label $y_{\text{pre}}^{\text{stu}}$ and the manual label $y$ of the image $X$, respectively.

\section{DTN Architecture}\label{sec_architecture}
Besides the CNN-based backbone, the proposed FFD model DTN consists of three key modules: the mixture of expert (MoE) module, the locally-enhanced vision transformer (LEVT) module, and the multi-attention scaling (MAS) method. 
As Figure~\ref{fig3} shows, given an input facial image $X$, DTN first extracts the local fine feature $X_{\text{loc}}$ through a CNN-based backbone,
which is then fed into the MoE module to capture diverse robust forgery embeddings $X_\text{MoE}$. 
After that, $X_\text{MoE}$ is transferred into LEVT integrated with MAS, 
to generate comprehensive forgery representations $f$, where richer facial forgery patterns are explored by adaptively choosing attention maps, and locally-enhanced global relations among image patches are explored.
Finally, a classifier consisting of a full connection layer transforms the diverse locally-enhanced global pattern $f$ into the final predictions $y_\text{pre}$, i.e.,
\begin{align}
	\text{DTN}(X) &= \text{Cls}\circ\text{GAP}\circ\text{LEVT}\circ\text{MoE}\circ\text{BaB}(X) \nonumber \\
	&= \text{Cls}\circ\text{GAP}\circ\text{LEVT}\circ\text{MoE}(X_{\text{loc}})\nonumber \\
	&= \text{Cls}\circ\text{GAP}\circ\text{LEVT}(X_\text{MoE})\nonumber \\
	&= \text{Cls}\circ\text{GAP}(X_{\text{Tra}})\nonumber \\
	&= \text{Cls}(f) = y_{\text{pre}},
\end{align}
where $\circ$ denotes the function composition, ``BaB'' denotes the CNN-based backbone, 
and ``Cls'' and ``GAP'' are a classifier and a global average pooling, respectively.

In our work, we take VGG as the CNN-based backbone (BaB), which initially extracts local fine features from the input facial image $X$,
\begin{align}
	X_{\text{loc}} = \text{BaB}(X).
\end{align}
In the following of this section, we will provide details of MoE and LEVT used in DTN.

\subsection{Mixture of Expert Module}\label{subsec31}
In DTN, we devise the mixture of expert (MoE) module to explore the various robust forgery patterns from $X_{\text{loc}}$. Unlike vanilla MoE \citep{moe} which adjusts model capacity using anchored triggered settings for analyzing data, our MoE module introduces the random Gaussian noise and flexibly integrates the decisions of different experts.

Specifically, given the local forgery feature $X_{\text{loc}}\in\mathbb R^{c\times h\times w}$ with channel $c$, height $h$, and width $w$,
MoE first add it with $B$ random Gaussian noise data $\big\{N_i^{\text{gau}}\in\mathbb R^{c\times h\times w}\big\}_{i=1}^B$,  respectively, then adopts $B$ expert blocks to mine diverse robust forgery traces, i.e.,
\begin{align}\label{eq_moe}
	M_i & = \text{Ex}_i \big(X_{\text{loc}}+N_i^{\text{gau}}\big)\in\mathbb R^{1\times h\times w},
\end{align}
where $\text{Ex}_i$ is the $i$-th expert block, $i=1,2,\cdots,B$, which consists of two convolutional layers with kernel size of $1\times1$ and a sigmoid function, generating the attention map $M_i$ with one channel.

To facilitate the flexible integration of expert decisions, MoE employs the MAS method to modulate the $B$ attention maps using $B$ learnable scale factors $\{\theta_i\}_{i=1}^B$, 
which are then concatenated together and applied with max-pooling to generate diverse attention map $A\in\mathbb R^{h\times w}$,
\begin{align}
	A = \sigma\bigg(\text{MaxP}\bigg(\text{Cat}\Big(\Big\{\sigma(\theta_i)\cdot M_i\Big\}_{i=1}^B\Big)\bigg)\bigg),
\end{align}
where $\sigma$ is the sigmoid function to confine the learnable factors and the final attention coefficients in the interval of $[0,1]$,
``Cat'' denotes the concatenation operation, and ``MaxP'' is the max-pooling along the channel direction.

Finally, MoE generates the robust forgery patterns $X_\textrm{MoE}$ using a broadcast element-wise multiplication,
\begin{gather}
	X_\textrm{MoE}=X\odot A \in\mathbb{R}^{c\times h\times w},
\end{gather}
which is then fed into the LEVT.

\begin{figure*}[t!]
	\centering
	\includegraphics[width=\linewidth]{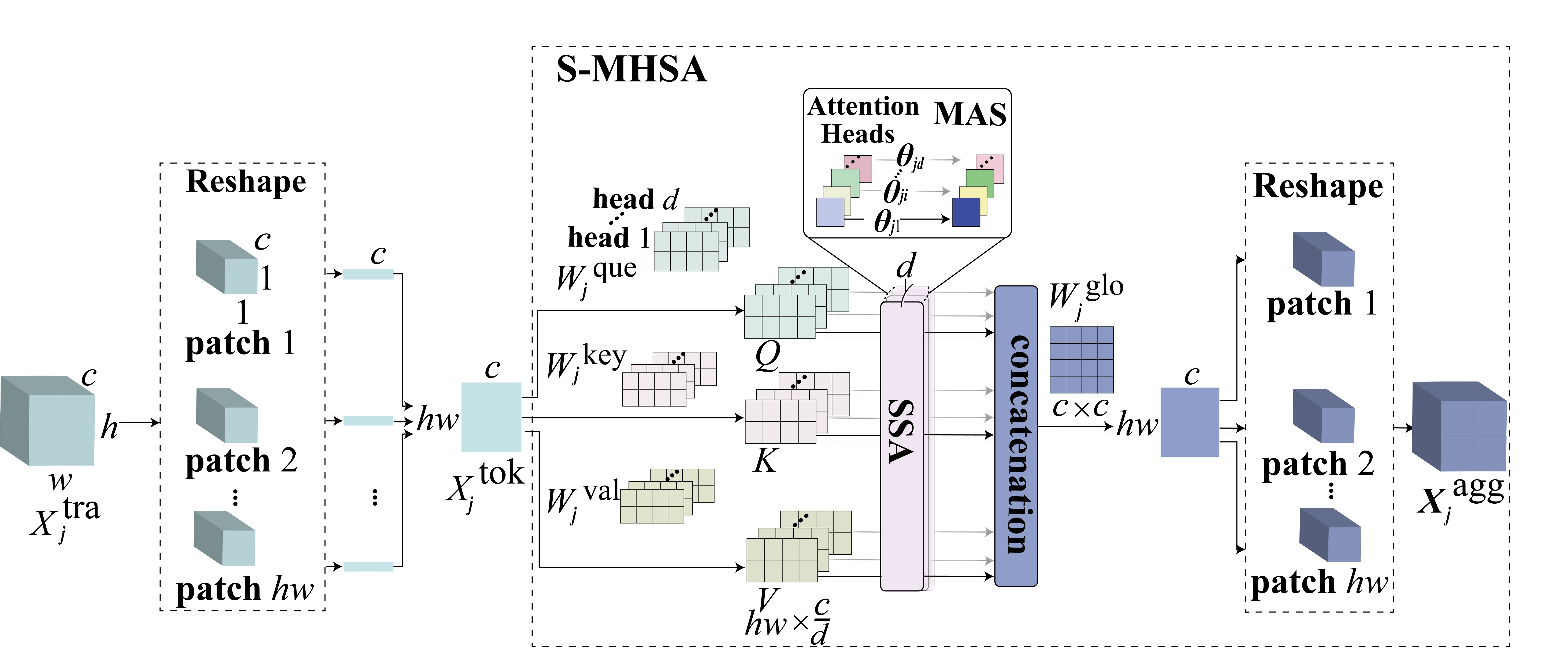} 
	\captionsetup{labelsep=none}
	\caption{ The workflow of the scaling multi-head self-attention (S-MHSA) module.}\label{fig311}
\end{figure*}

\subsection{Locally-enhanced Vision Transformer}\label{subsec33}
To capture comprehensive locally-enhanced global representations in the robust feature $X_{\text{MoE}}$, we design locally-enhanced vision transformer (LEVT), using a scaling multi-head self-attention (S-MHSA) mechanism with MAS, and local connection (LC). 

In detail, as Figure~\ref{fig3} shows, LEVT contains $L$ stacked transformer blocks, $\textrm{TB}_j,j=1,2,...,L$.
Before being input into LEVT, the feature $X_{\text{MoE}}$ is first added with a learnable positional embedding $P\in\mathbb R^{c\times h\times w}$ to encode positional information
\begin{align}
	X_1^{\text{tra}} & = X_{\text{MoE}} + P,
\end{align}
which is then sequentially fed into $L$ blocks, i.e.,
\begin{align}
	\label{eq_levt}
	\textrm{LEVT}(X_1^{\text{tra}}) &= \textrm{TB}_{L} \circ \textrm{TB}_{L-1} \circ \cdots \circ \textrm{TB}_{2} \circ \textrm{TB}_{1}(X_1^{\text{tra}}) \nonumber \\
	&= \textrm{TB}_{L} \circ \textrm{TB}_{L-1} \circ \cdots \circ \textrm{TB}_{2}(X_2^{\text{tra}}) \nonumber \\
	&= \cdots \nonumber\\
	&= \textrm{TB}_{L} \circ \textrm{TB}_{L-1}(X_{L-1}^{\text{tra}}) \nonumber \\
	&= \textrm{TB}_{L}(X_L^{\text{tra}}) = X_{\text{Tra}}.
\end{align}

{\bfseries\setlength\parindent{1.3em} Transformer block.} Specifically, as Figure~\ref{fig3} shows, in the $j$-th Transformer block TB$_j$, 
a batch normalization (BN) is first applied on the feature $X_j^{\text{tra}}$ to equalize the feature among the channels and facilitate the model learning.
As Figure~\ref{fig311} illustrates, $X_j^{\text{tra}}$ is then reshaped into a 2D tensor by flattening feature vectors along channels,
\begin{align}
	X_j^{\text{tok}} = & \text{Reshape}(\text{BN}(X_j^{\text{tra}})) \in \mathbb R^{hw\times c},
\end{align}
i.e., $X_j^{\text{tok}}$ takes as its row (token) every vector with length of $c$ at each spatial position in the normalized $X_j^{\text{tra}}$.  
We then perform scaling multi-head self-attention mechanism (S-MHSA) on $X_j^{\text{tok}}$ to learn diverse global forgery representations.  Previous approaches \cite{UIA-ViT,wodajo2021deepfake} solely mine global features using MHSA, which tends to struggle to explore diverse global relations among patches due to attention collapse. Therefore, we propose to enhance the diversity of global features via S-MHSA.

As Figure~\ref{fig311} shows, in S-MHSA, $X_j^{\text{tok}}$ is evenly partitioned into $d$ heads,
\begin{align}
	\big\{X_{j,i}^{\text{tok}}\in\mathbb R^{hw\times \frac{c}{d}}\big\}_{i=1}^d = \text{Pa}\big(X_j^{\text{tok}}\big) \in \mathbb R^{d\times hw\times \frac{c}{d}},
\end{align} 
where ``Pa'' is a partition operation along the row direction. 
Then, a scaling self-attention (SSA) mechanism is conducted on each head,
which first computes query, key, and value matrices using three learnable weight matrices $W_{j,i}^{\text{que}}, W_{j,i}^{\text{key}}$, and $W_{j,i}^{\text{val}}\in\mathbb{R}^{\frac{c}{d}\times \frac{c}{d}}$, respectively.
\begin{align}
	Q_{j,i}=& X_{j,i}^{\text{tok}}\,W_{j,i}^{\text{que}},\\
	K_{j,i}=& X_{j,i}^{\text{tok}}\,W_{j,i}^{\text{key}},\\
	V_{j,i}=& X_{j,i}^{\text{tok}}\,W_{j,i}^{\text{val}},
\end{align}
then computes the attention weight using MAS operation to mitigate attention collapse,
\begin{align}
	A_{j,i} = \delta\bigg(\sigma(\theta_{j,i}) \odot \frac{Q_{j,i} K_{j,i}^T}{\sqrt{c/d}}\bigg),
\end{align} 
where $\theta_{j,i}\in\mathbb R^{1\times1}$ is a learnable scale factor and $\sigma$ is a sigmoid function to confine its value in the interval of $[0,1]$,
and $\delta$ is softmax function.

Finally, the self-attention outputs from all heads are concatenated to yield global forgery features,
\begin{align}
	X_j^{\text{glo}} = \text{Cat}\Big(\big\{A_{j,i}V_{j,i}\big\}_{i=1}^d\Big) \in\mathbb R^{hw\times c},
\end{align}
which is then aggregated using a weight matrix with a size of $c\times c$ and reshaped to a 3D tensor,
\begin{align}
	X_j^{\text{agg}} = \text{Reshape}(X_j^{\text{glo}} W_j^{\text{glo}})\in\mathbb R^{c\times h\times w}.
\end{align}

After S-MHSA, TB$_j$ applies a local connection (LC) layer consisting of a convolutional layer to produce the locally-enhanced global representation. As Fan et al. \citep{fan-etal-2021-mask} demonstrate, in the traditional MHSA module, each feature patch is equally accessible to any other ones, and feature patches not in the neighborhood may also attend to each other with relatively large scores. Such a problem is likely to add noises to semantic modeling and ignore the link among surrounding signals. Consequently, we propose a local connection layer to refine global embeddings, and thus explore locally-enhanced global representations. Then TB$_j$ conducts the residual addition operation,
\begin{align}
	X_j^{\text{lc}} = \text{LC}(X_j^{\text{agg}}) + X_j^{\text{agg}}.
\end{align}
Thereafter, TB$_j$ finally outputs $X_{j+1}^{\text{tra}}$ using a residual module integrated with a BN operation and a multi-layer perceptron (MLP). Unlike the traditional MLP, the fully connected layer is replaced by $1\times1$ convolution to reduce parameters and computational overheads.
\begin{align}
	X_{j+1}^{\text{tra}} = \text{MLP}_j \circ \text{BN}(X_j^{\text{lc}}) + X_j^{\text{lc}}.
\end{align}

As Eq. (\ref{eq_levt}) shows, the last one TB block outputs $X_{\text{Tra}}$, which is transformed into the feature $f\in\mathbb R^{1\times c}$ of the input facial image $X$, by a global average pooling (GAP),
and then used to compute the predicted label by a classifier, i.e.,
\begin{align}
	y_{\text{pre}} = \text{Cls}(f) = \text{Cls}\circ\text{GAP}(X_{\text{Tra}}).
\end{align}

\section{Experiments}\label{sec4}
\subsection{Datasets}\label{subsec41}

We employed five benchmark databases to evaluate the robustness of our model: FaceForensics++ (FF++) \citep{Rossler2019FaceForensics}, Deepfake Detection Challenge (DFDC) \citep{Brian2020The}, Celeb-DF \citep{Li2020Celeb-DF}, DeeperForensics-1.0 (DF-1.0) \citep{Jiang2020DeeperForensics}, and Deepfake Detection Dataset (DFD). FF++ consists of videos with various compression levels, namely raw, high quality (HQ), and low quality (LQ). We utilized HQ videos along with the official splits, i.e. 720 videos for training, 140 videos for validation, and 140 videos for testing.

\subsection{Implementation Details}\label{subsec42}

We developed the proposed approach in the open-source PyTorch framework and performed all the experiments on the Tesla V100 GPU with batch size 32. We augmented the original frames 4 times for category balance in the training process. We utilized dlib \citep{dlib} to crop face regions as input images with a size of 256$\times$256. The number of branches $B$ in MoE is set to 2. The depth $L$ of transformer blocks is set to 6 and the attention heads $d$ are set to 8. The Albumentations \citep{info11020125} library is exploited for data augmentation. Our model is trained with the Adam optimizer with a learning rate of 1e-4 and weight decay of 1e-4. We used the scheduler to drop the learning rate by ten times every 15 epochs. The balance weights $\alpha_1$, $\alpha_2$, and $\alpha_3$ in Eq.~(\ref{eq7}) are set to 4, 0.4, and 0.6, respectively.

{\bfseries\setlength\parindent{0em}Evaluation Metrics.} We used accuracy (ACC) and area under the receiver operating characteristic curve (AUC) as our evaluation metrics.
\begin{table*}[t]
	\captionsetup{labelsep=none}
	\caption{\ Within-dataset and cross-dataset evaluation. ACC and AUC scores (\%) on Celeb-DF, DFDC, and DF-1.0, after training on FF++. \textsuperscript{\textdagger} denotes the results of the models are reproduced by ourselves. The best results are in bold\label{tab1}}
	\small
	\setlength{\tabcolsep}{0.5mm}{
		\begin{tabular*}{\textwidth}{@{\extracolsep{\fill}}lcccccccccc}
			\toprule
			\multirow{2}*{Method} & \multicolumn{2}{c}{FF++} & \multicolumn{2}{c}{Celeb-DF} & \multicolumn{2}{c}{DFDC} & \multicolumn{2}{c}{DF-1.0} & \multicolumn{2}{c}{DFD} \\
			\cmidrule{2-4}\cmidrule{4-6}\cmidrule{6-7}\cmidrule{7-9}\cmidrule{9-11}
			& ACC   & AUC   & ACC   & AUC   & ACC   & AUC   & ACC   & AUC   & ACC   & AUC \\
			\midrule 
			CViT\textsuperscript{\textdagger}\citep{wodajo2021deepfake}  & 90.47 & 96.69 & 50.75 & 64.70 & 60.95 & 65.96 & 56.15 & 62.42 & 77.70 & 89.28 \\
			Xception\citep{Rossler2019FaceForensics} & 95.73& 96.30 & - & - & - & - & - & - & - & - \\
			TwoStream\citep{luo2021generalizing} & - & 95.05 & -& 79.40& - & 79.70 & -& 73.80& - & 91.90\\
			SBIs\citep{Shiohara_2022_CVPR}  & - & 99.64 & - & 93.18& - & 72.42 & - & - & - & 97.56 \\
			EfficientViT\textsuperscript{\textdagger}\citep{coccomini2022combining} & 88.26 & 96.14 & 49.00 & 62.47 & 64.78 & 70.12 & 66.67 & 70.60 & 80.75 & 90.50 \\
			CrossEfficientViT\textsuperscript{\textdagger}\citep{coccomini2022combining} &93.67 & 98.36 & 44.24 & 65.29 & 66.14 & 75.55 & 62.16 & 67.51 & 86.71 & 94.23 \\
			RECCE\citep{cao2022end} & 97.06 & 99.32 & - & 68.71 & - & 69.06 & - & - & - & - \\
			IID\citep{IID} & - & 99.32 & - & 83.80 & - & \textbf{81.23} & - & - & - & 93.92 \\
			IDDDM\citep{IDDDM} & - & \textbf{99.70} & - & 91.15 & - & 71.49 & - & - & - & - \\
			UIA-ViT\citep{UIA-ViT} & - &99.33 & - &82.41 & - & 75.80& - & - & - &94.68 \\
			FoCus\citep{FoCus} & 96.43 &99.15&  &76.13 & - & 68.42& - & - & - &- \\
			ResNet34\citep{ed2024} & - &98.30&  &86.40 & - & 72.10& - & - & - &- \\
			Yu et al. \citep{yu}& \textbf{97.72} &99.55&  -&72.86& -& 69.23&-& - & - &80.76 \\
		Guan et al. \citep{Guan}& - &99.17&  -&\textbf{95.14}& - &74.65& - & - & - &-\\
			NACO \citep{LearningNC} & - &99.70&  -&89.50& -& 76.70& -& - & - &-\\
			DTN (Ours)  & 97.13 & \textbf{99.70} & \textbf{70.68} & 75.32 & \textbf{72.13}  &  80.01& \textbf{71.24} &  \textbf{78.77}& \textbf{87.32} & \textbf{97.60} \\
			\midrule 
		\end{tabular*}	
	}
\end{table*}

\subsection{Comparison with the State of the Art}\label{subsec44}

{\bfseries\setlength\parindent{0em} Within-dataset evaluation.} 
To fully evaluate the effectiveness of the proposed network, we performed within-dataset evaluations. We trained models using FF++, and tested them on the same dataset. As Table~\ref{tab1} displays, our model outperforms most approaches on FF++. For transformer-based models, the AUC of our network is around 3\%, 3.6\%, and 1.4\% higher than that of CViT, EfficientViT, and CrossEfficientViT, respectively, which attributes to the powerful modeling capabilities of our model. 	

{\bfseries\setlength\parindent{0em} Cross-dataset generalization.} 
To fully validate the effectiveness of the proposed framework, we conducted extensive cross-dataset evaluations. We tested models on FF++, DFDC, Celeb-DF, DF-1.0, and DFD after training on FF++. As Table~\ref{tab1} shows, our model surpasses most methods by a wide gap. It also exceeds the existing state-of-the-art approach, RECCE, by 6.6\% AUC on Celeb-DF. Unlike RECCE which captures local information, we argue that our method can learn richer local and global features, to discover diverse traces of forgery faces. Compared to the transformer-based model CViT which considers both local and global features, our method shows that it is valuable to study various robust features and locally enhanced global representations. Especially for the DF-1.0 dataset, it is a challenging benchmark since extensive real-world perturbations are applied. The AUC of our method is about 8.1\% and 11.2\% higher than that of  EfficientViT and CrossEfficientViT, respectively, achieving remarkable performance on DF-1.0. Compared to the latest detectors, the AUC of our model on DFDC is 3.3\%, 5.4\%, and 11.8\% higher than that of \citep{LearningNC}, \citep{Guan} and \citep{yu}, respectively, demonstrating the effectiveness and generalization of our model. These results show that the proposed approach is more general to various deepfake datasets than existing methods. 

\begin{table*}[t!]
	\captionsetup{labelsep=none}
	\caption{\ Cross-manipulation generalization\label{tab10}}	
	\small
	\setlength{\tabcolsep}{0.0mm}{
		\begin{tabular*}{\textwidth}{@{\extracolsep{\fill}}lccccccc}
			\toprule
			\multirow{2}[4]{*}{Training Set} & \multirow{2}[4]{*}{Model} & \multicolumn{6}{c}{Testing Set (AUC)} \\
			\cmidrule{3-5} \cmidrule{5-8}      &       & DF    & F2F   & FS    & NT & \\
			\midrule 
			\multirow{4}[1]{*}{DF}  & Xception \citep{Rossler2019FaceForensics} & 99.38 & 75.05 & 49.13 & 80.39\\ 
			& Face X-ray \citep{Li_2020_CVPR}  & 99.17 & 94.14 & \textbf{75.34} & 93.95 \\
			& SOLA \citep{fei2022learning} & 100   & 96.95 & 69.72 & 98.48 \\
			& IID \citep{IID} & 99.51  & - & 63.83 &- \\
			& IDDDM \citep{IDDDM} & 100  & 83.94 & 58.33 &68.98 \\
			& Guan et al. \citep{Guan} & 100  & 66.72 & 43.29&67.40 \\
			& DTN (ours) &   \textbf{100}   &  \textbf{97.23}     &  74.26     & \textbf{98.53}  \\
			\midrule
			\multirow{4}[1]{*}{F2F} & Xception \citep{Rossler2019FaceForensics}  & 87.56 & 99.53 & 65.23 & 65.90 \\
			& Face X-ray \citep{Li_2020_CVPR}& 98.52 & 99.06 & 72.69 & 91.49 \\
			& SOLA  \citep{fei2022learning} & 99.73 & 99.56 & 93.50 & 96.02 \\
			& IID \citep{IID} & -  & - & - &- \\
			& IDDDM \citep{IDDDM} & 99.88  & 99.97 & 79.40 &82.38 \\
				& Guan et al. \citep{Guan} & 89.10  & \textbf{100} & 62.51&48.61 \\
			& DTN (ours) &   \textbf{100}    & 99.82      &  \textbf{93.63}     & \textbf{97.03}  \\
			\midrule
			\multirow{4}[2]{*}{FS} & Xception \citep{Rossler2019FaceForensics} & 70.12 & 61.7  & 99.36 & 68.71\\
			& Face X-ray \citep{Li_2020_CVPR}& 93.77 & 92.29 & 99.20  & 86.63 \\
			& SOLA \citep{fei2022learning} & 99.11 & 98.13 & \textbf{ 99.98 }  & 92.07 \\
			& IID \citep{IID} & 75.39  & 99.73 &66.18 &80.43 \\
			& IDDDM \citep{IDDDM} & 93.42  & 74.00 & 99.92 &49.86 \\
				& Guan et al. \citep{Guan} & 64.19 & 66.03& 99.97&37.37 \\
			& DTN (ours) &  \textbf{ 99.15 }     &  \textbf{ 99.74 }     &   98.21    & \textbf{ 92.45}  \\
			\midrule
			\multirow{4}[2]{*}{NT} & Xception \citep{Rossler2019FaceForensics} & 93.09 & 84.82 & 47.98 & 99.50 \\
			& Face X-ray  \citep{Li_2020_CVPR} & 99.14 & \textbf{98.43} & 70.56 & 98.93 \\
			& SOLA \citep{fei2022learning}  & 99.64 & 97.69 & 90.20 & 99.76 \\
			& IID \citep{IID} &-  & - &- &- \\
			& IDDDM \citep{IDDDM} & \textbf{100}  & 97.93 & 86.76 &99.46 \\
			& Guan et al. \citep{Guan} & 90.15 & 68.83& 38.28&97.84 \\
			& DTN (ours) & 99.71    &  97.95    &  \textbf{ 91.25}    &  \textbf{99.81}\\
			\bottomrule
		\end{tabular*}%
	}
	
\end{table*}%

{\bfseries\setlength\parindent{0em} Generalization performance to novel forgery.} We trained models using one of the four
forgeries in FF++ and tested the trained models on the remaining three. As Table ~\ref{tab10} shows, DTN achieves promising generalization to unseen face forgery, exceeding most methods by a large margin. The AUC of DTN
is the best among all approaches, when F2F and NT are used as the training set. When DF and FS are used as the training set, the AUC of our approach ranks top two, a little bit lower than Face X-ray and SOLA, respectively. As Face Swap generates more unique forgery artifacts than other deep learning-based manipulations, i.e.DF, F2F, and NT, and Face X-ray is designed to focus on the blending boundaries, Face X-ray  achieves better performance in detecting
DF and FS manipulations, than our approach.
\begin{figure*}[t!]%
	\includegraphics[width=\textwidth]{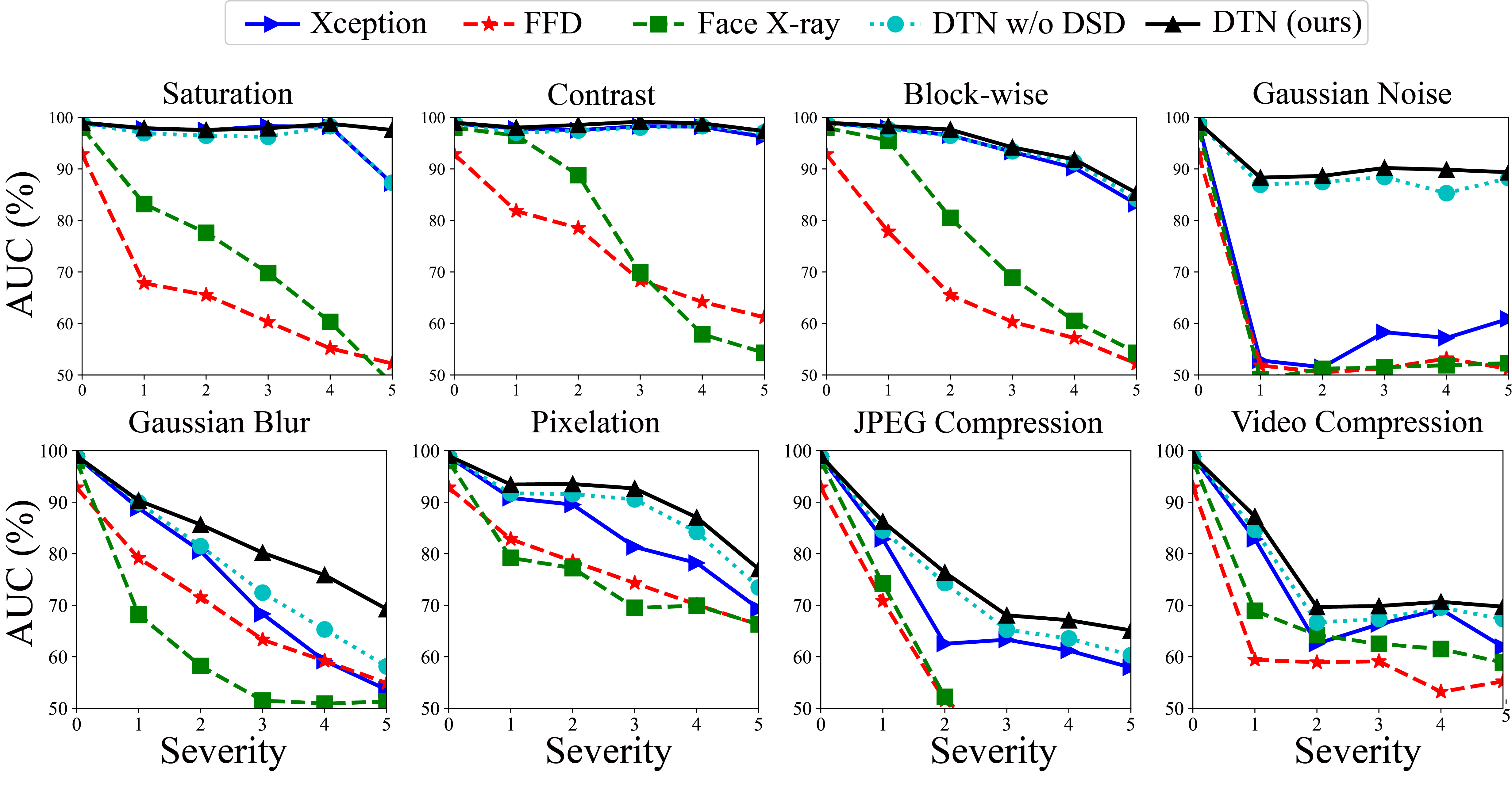}
	\centering
	\captionsetup{labelsep=none}
	\caption{\ Robustness to unseen image deformations. DSD denotes deepfake self-distillation}\label{fig21}
\end{figure*}

{\bfseries\setlength\parindent{0em} Robustness to common image corruptions.} With the widespread usage of image manipulation techniques on social media platforms, it is essential that face forgery detection systems are robust against typical perturbations. In this experiment, we aimed to investigate the robustness of the detectors against various unseen corruptions, by training them on FF++ (HQ) and then evaluating their performance on distorted samples \citep{Jiang2020DeeperForensics}. Like \citep{Li_2020_CVPR}, various image corruptions with a range of severity levels, including saturation adjustments, contrast modifications, block-wise distortions, white Gaussian noise addition, blurring, pixelation, and video compression using the H.264 codec, are tested. A severity of 0 means that no degradation is applied. Specifically, JPEG compression is achieved by adjusting the compression quality parameters [20, 25, 30, 45, 40] (The larger the parameter value, the greater the compression rate). Similarly, video compression controls the compression quality of the video via the constant rate factor [30, 32, 35, 38, 40] (The larger the parameter value, the greater the compression rate). As Gaussian noise has been used in the MoE module of our DTN, we substitute the Gaussian noise with the Uniform noise to address the data leakage problem. Figure~\ref{fig21} illustrates the variations of AUC for all approaches, when corruptions with different intensities are applied. One can observe from the figures that DTN consistently performs better than all of the other approaches, across all of the different image corruptions. Xception is highly susceptible to various deformations, primarily due to the extraction of local forgery representations via heavy dependence on a restricted receptive domain.

\subsection{Ablation Study}\label{subsec43}

{\bfseries\setlength\parindent{0em} Impacts of components.}
In order to evaluate the contribution of each module to learning ability, we observed the performance on FF++ and DF-1.0 after training on FF++. The results of model ablation are shown in Table~\ref{tab2}, which shows the effect of gradually adding LEVT, MoE, and MAS components to the baseline network.

\begin{table}
	\centering
	\captionsetup{labelsep=none}
	\caption{\ Model ablation. The ACC and AUC scores (\%) on FF++ and DF-1.0 after training on FF++. Baseline means the CNN backbone VGG. * denotes the module with MAS. The best results are in bold \label{tab2}}
	\small
	\begin{tabular}{@{\extracolsep\fill}ccccc@{}}
		\toprule
		\multirow{2}*{Method} & \multicolumn{2}{c}{FF++} & \multicolumn{2}{c}{DF-1.0} \\  \cmidrule{2-3} \cmidrule{4-5}
		& ACC & AUC & ACC & AUC \\
		\midrule
		Baseline            & 91.97 & 97.42 & 60.56 & 68.36 \\
		Baseline+LEVT        & 93.03 & 98.02 & 62.86 & 70.88 \\
		Baseline+LEVT+MoE     & 94.76 & 98.93 & 65.45 & 73.73 \\
		Baseline+LEVT+MoE*    & 95.02 & 99.02 & 65.83 & 74.57 \\
		Baseline+LEVT*+MoE    & 95.56 & 99.30 & 66.96 & 75.86 \\
		Baseline+LEVT+MoE+MAS & \textbf{97.13} & \textbf{99.70} & \textbf{71.24} & \textbf{78.77} \\
		\bottomrule
	\end{tabular}
\end{table} 

 In comparison to the baseline, LEVT increases the performance by 1.0\% ACC and 0.6\% AUC, confirming the importance of locally-enhanced global embeddings. The gains from introducing the MoE module (+0.9\%) are slightly obvious, demonstrating that diverse local features offer useful information to benefit deepfake detection. We observed that MAS improves performance by 2.3\% ACC and 0.8\% AUC. We believed that several attention heads are adaptively emphasized and suppressed, respectively, guiding the model to learn rich and discriminative forgery regions. Our method finally achieves 97.13\% ACC and 99.70\% AUC on FF++, an increase of 5.1\% ACC and 2.3\% AUC, respectively, when compared to the baseline.

{\bfseries\setlength\parindent{0em} Impacts of STG.} In order to evaluate the contribution of the STG module to learning ability, we observed the performance on FF++, Celeb-DF, DFDC and DF-1.0 after training on FF++. The results of model ablation are shown in Table~\ref{tab38}. The gains from introducing the STG module (+3.4\%) are evident, showing that soft labels offer valuable information to facilitate deepfake detection.

\begin{table*}[t!]
	\captionsetup{labelsep=none}
	\caption{\ Ablation of STG. The best results are in bold\label{tab38}}
	\small
	\begin{tabular*}{\textwidth}{@{\extracolsep\fill}ccccccccc@{}}
		\toprule
		\multirow{2}*{Loss function} & \multicolumn{2}{c}{FF++} & \multicolumn{2}{c}{\hspace{0.8em}Celeb-DF} & \multicolumn{2}{c}{DFDC} & \multicolumn{2}{c}{DF-1.0} \\
		\cmidrule{2-4} \cmidrule{4-5} \cmidrule{5-7} \cmidrule{7-9}
		& ACC & AUC & ACC & AUC & ACC & AUC & ACC & AUC \\
		\midrule
		
		DTN w/o STG & 95.20 & 99.17 & 64.23 & 73.72 & 70.39 & 78.01 & 68.03 & 75.36 \\
		DTN w/ STG & \textbf{97.13} & \textbf{99.70} & \textbf{66.68} & \textbf{75.32} & \textbf{72.13} & \textbf{80.01} & \textbf{71.24} & \textbf{78.77} \\
		\bottomrule
	\end{tabular*}
\end{table*}



{\bfseries\setlength\parindent{0em} Influences of MoE.} To verify the efficiency of MoE, we compared it with the vanilla MoE. The results are shown in Table~\ref{tab29}. Our MoE method shows excellent performance on various datasets, showing the effectiveness of adaptively incorporating the information of different experts. 

\begin{figure}[t]%
	\includegraphics[width=\linewidth]{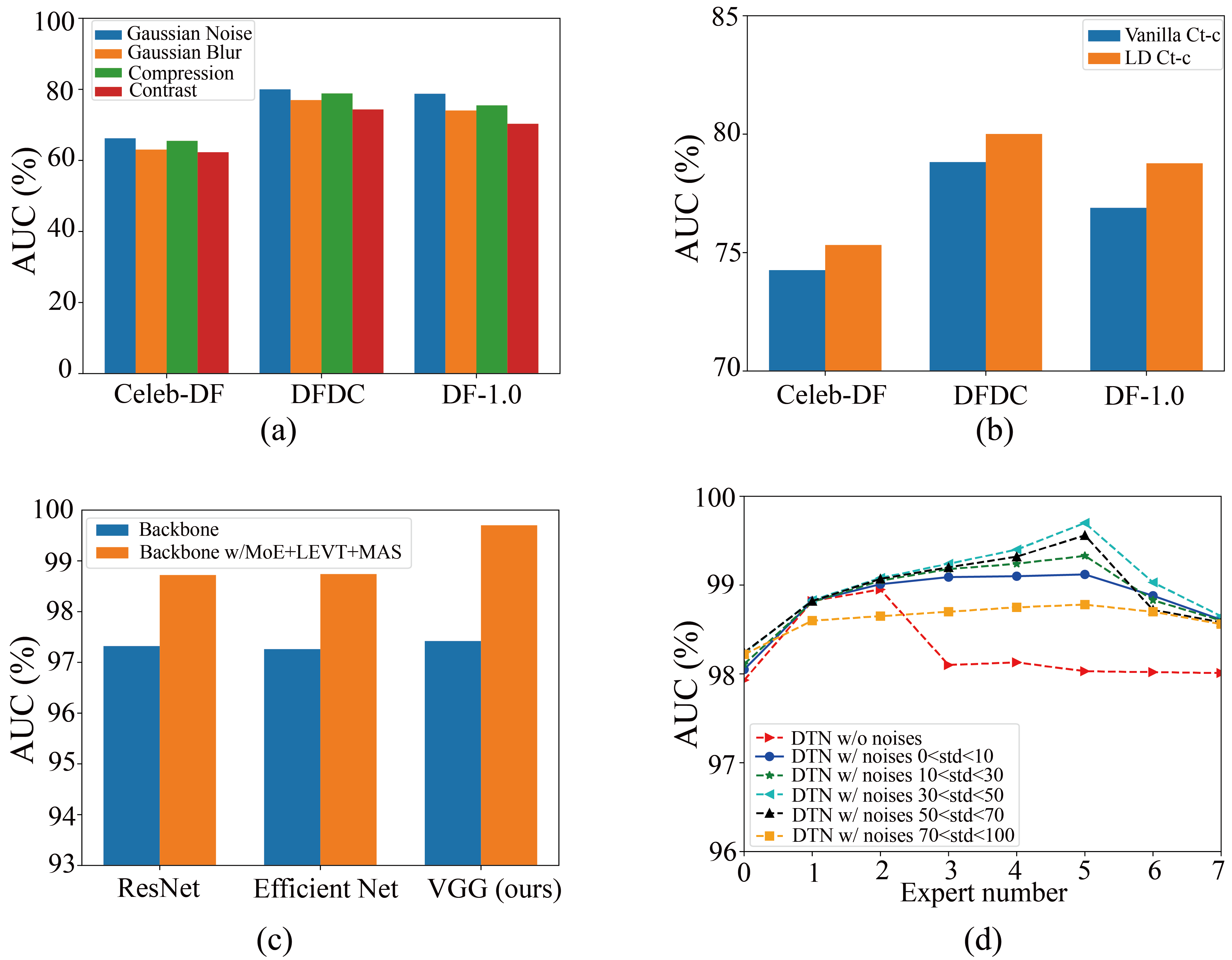}
	\centering
	\captionsetup{labelsep=none}
	\caption{ \ (a) The impact of various augmentation technologies in MoE. (b) The effect of the low dimensional (LD) Ct-c loss. (c) Generalization to different backbones. (d) The effect of random Gaussian noises and the number of experts in MoE.}\label{fig28}
\end{figure}

{\bfseries\setlength\parindent{0em} Impact of various augmentation technologies in MoE.} We studied the effect of various augmentation techniques in MoE such as Gaussian noise, Gaussian blur, compression, and contrast. We trained models using the same level of augmentations on FF++, and tested them on Celeb-DF, DFDC and DF-1.0. As shown in Figure~\ref{fig28} (a), the AUC with Gaussian noises attains maximum, compared to other augmentation methods, indicating that noises play an important role in face forgery detection tasks.

{\bfseries\setlength\parindent{0em} Impacts of loss functions.} To demonstrate the contribution of the proposed loss function, we conducted experiments on different losses. The results are shown in Table~\ref{tab8}. When the network is supervised with only cross-entropy loss, the AUC is 67.75\% on DF-1.0. However, a nearly 8\% growth of AUC could be achieved by introducing the Ct-c loss, verifying the importance of intra-class compactness and inter-class separability. Meanwhile, we noticed that our deepfake self-distillation loss shows the best among these losses, which suggests that promising results can be acquired by combining metric learning with knowledge distillation.We further performed experiments to verify the effect of low dimensional Ct-c loss. Models are trained with vanilla Ct-c loss and low dimensional Ct-c loss, respectively, on FF++, and tested on Celeb-DF, DFDC, and DF-1.0. As Figure~\ref{fig28} (b) shows, models show better generalization performance when supervised using the low dimensional Ct-c loss.
\begin{table*}[t!]
	\captionsetup{labelsep=none}
	\caption{\ Ablation of various losses. CE and Ct-c denote the cross-entropy loss and the contrastive-center loss, respectively. The best results are in bold\label{tab8}}
	\small
	\begin{tabular*}{\textwidth}{@{\extracolsep\fill}ccccccccc@{}}
		\toprule
		\multirow{2}*{Loss function} & \multicolumn{2}{c}{FF++} & \multicolumn{2}{c}{Celeb-DF} & \multicolumn{2}{c}{DFDC} & \multicolumn{2}{c}{DF-1.0} \\
		\cmidrule{2-4} \cmidrule{4-5} \cmidrule{5-7} \cmidrule{7-9}
		& ACC & AUC & ACC & AUC & ACC & AUC & ACC & AUC \\
		\midrule
		CE loss & 93.03 & 98.13 & 57.84 & 70.21 & 67.94 & 73.32 & 65.89 & 67.75 \\
		KD + CE loss & 95.77 & 99.30 & 62.72 & 70.92 & 69.83 & 77.95 & 66.97 & 71.14 \\
		Ct-c + CE loss & 95.20 & 99.17 & 64.23 & 73.72 & 70.39 & 78.01 & 68.03 & 75.36 \\
		DSD loss & \textbf{97.13} & \textbf{99.70} & \textbf{66.68} & \textbf{75.32} & \textbf{72.13} & \textbf{80.01} & \textbf{71.24} & \textbf{78.77} \\
		\bottomrule
	\end{tabular*}
\end{table*}

{\bfseries\setlength\parindent{0em} Generalization to various backbones.} To evaluate the generalizability of the model across different CNN backbones, we selected several common backbones such as ResNet and EfficientNet, and combined them with our MoE, LEVT, and MAS modules for testing. We trained models using FF++ and tested them on the same dataset. As Figure~\ref{fig28} (c) shows, due to the introduction of our modules, the backbones consistently exhibit better performance.

{\bfseries\setlength\parindent{0em} Effect of random noises and the number of experts in MoE.} We further investigated the influence of Gaussian noises and the number of experts in MoE, when our DTN was trained using FF++ (HQ). We reported the performance of DTN from 1 to 7, when Gaussian noise is added or not. As Figure~\ref{fig28} (d) shows, the performance generally improves with the increase of expert number, when Gaussian noise is involved. The AUC achieves the maximum when five experts are used and starts to decrease thereafter. By contrast, the AUC without involving Gaussian noises reaches maximum when two experts are used and then stays a stable performance when the number of experts equals or is larger than three. We believe that generally it is easier to capture more forgery regions, when more experts are used. The addition of Gaussian noises may further help capture richer forgery patterns to enhance the robustness of the network.

\begin{table*}[t!]
	\centering
	\captionsetup{labelsep=none}
	\caption{\ The ablation of the local connection layer. We tested models on FF++, Celeb-DF, DFDC, and DF-1.0, after training on FF++. LC denotes the local connection layer. Best results are in bold\label{tab23}}
	\small
	\setlength{\tabcolsep}{0.5mm}{%
		\begin{tabular*}{\textwidth}{@{\extracolsep\fill}lccccccccccccccc}
			\toprule
			\multirow{2}*{Method} & \multicolumn{2}{c}{FF++} & \multicolumn{2}{c}{Celeb-DF} & \multicolumn{2}{c}{DFDC} & \multicolumn{2}{c}{DF-1.0}& \multicolumn{1}{c}{ Params (M)}     &  \multicolumn{1}{c}{FLOPs (G)}  \\
			
			\cmidrule(lr){2-4} \cmidrule(lr){4-5} \cmidrule(lr){5-7} \cmidrule(lr){7-9}
			& ACC & AUC & ACC & AUC & ACC & AUC & ACC & AUC & & \\
			\midrule
			DTN w/o LC & 95.56 & 99.43 & 63.95 & 72.52 & 70.79 & 78.73 & 67.43 & 74.23 & 29.876 & 9.796 \\
			DTN  w/ LC & \textbf{97.13} & \textbf{99.70} & \textbf{66.68} & \textbf{75.32} & \textbf{72.13} & \textbf{80.01} & \textbf{71.24} & \textbf{78.77} & {29.912} & {9.799} \\
			\bottomrule
		\end{tabular*}%
	}
\end{table*}

\begin{table*}[t!]
	\centering
	\captionsetup{labelsep=none}
	\caption{\ The effect of the MoE module. We tested models on FF++, Celeb-DF, DFDC, DF-1.0 and DFD, after training on FF++. Best results are in bold\label{tab29}}
	\small
	\setlength{\tabcolsep}{1.5mm}{%
		\begin{tabular*}{\textwidth}{@{\extracolsep{\fill}}lcccccccccc}
			\toprule
			\multirow{2}*{Method} & \multicolumn{2}{c}{FF++} & \multicolumn{2}{c}{Celeb-DF} & \multicolumn{2}{c}{DFDC} & \multicolumn{2}{c}{DF-1.0} & \multicolumn{2}{c}{DFD} \\
			\cmidrule{2-4}\cmidrule{4-6}\cmidrule{6-7}\cmidrule{7-9}\cmidrule{9-11}
			& ACC   & AUC   & ACC   & AUC   & ACC   & AUC   & ACC   & AUC   & ACC   & AUC \\
			Vanilla MoE & 92.24 & 97.03 & 58.28 & 63.71 & 62.18 & 68.03 & 62.09 & 68.43 & 81.65 & 87.27 \\    Ous   & \textbf{93.21} & \textbf{98.20} & \textbf{60.23} & \textbf{66.22} & \textbf{64.88} & \textbf{72.67} & \textbf{65.45} & \textbf{74.36} & \textbf{82.93} & \textbf{90.58} \\    \bottomrule    \end{tabular*}%
	}
\end{table*}

{\bfseries\setlength\parindent{0em} Effect of local connection layers.} For the LEVT module, we designed the local connection layer to purify global representations. To study the importance of this component, we conducted ablation experiments in Table~\ref{tab23}. We observe that the local connection layer brings remarkable performance improvement. It turns out that local convolutional fusion on global features with latent noise, is very useful to extract locally enhanced global dependencies.

\begin{table*}[t!]
	\centering
	\captionsetup{labelsep=none}
	\caption{\ Ablation results of transformer-based models with MAS and with Re-attention. We tested models on FF++, Celeb-DF, DFDC, and DF-1.0, after training on FF++. EViT and CEViT denote the EfficientViT and CrossEfficientViT, respectively. The best results are in bold\label{tab3}}
	\small
		\setlength{\tabcolsep}{0.5mm}{%
		\begin{tabular*}{\textwidth}{@{\extracolsep\fill}lp{1.0cm}p{1.0cm}p{1.0cm}p{1.0cm}p{1.0cm}p{1.0cm}p{1.0cm}ccccccccccc}
			\toprule
			\multirow{2}*{Method} & \multicolumn{2}{c}{ FF++} & \multicolumn{2}{c}{\hspace{-0.8em}Celeb-DF} & \multicolumn{2}{c}{ DFDC} & \multicolumn{2}{c}{ DF-1.0}& \multirow{2}*{\shortstack{Params\\ (M)} } &  \multirow{2}*{\shortstack{FLOPs\\ (G)}}  \\
			\cmidrule(lr){2-4} \cmidrule(lr){4-5} \cmidrule(lr){5-7} \cmidrule(lr){7-9}
			& ACC   & AUC   & ACC   & AUC   & ACC   & AUC   & ACC   & AUC   & & \\
			\midrule
			ViT & 62.44 & 67.07 & 62.28 & 59.75 & 56.18 & 58.31 & 58.05 & 61.27 & 85.620 & 16.879 \\
			ViT w/ Re-att & 62.84 & 67.42 & 63.25 & 60.19 & 56.21& 58.34 & 58.21& 61.43 & 85.623 & 16.968 \\			
			ViT w/ MAS & 63.45 & 67.97 & 64.40 & 61.65 & 56.66 & 58.86 & 59.04 & 62.36 & 85.620 & 16.879 \\
			\midrule
			CViT & 90.47 & 96.69 & 50.75 & 64.70 & 60.95 & 65.96 & 56.15 & 62.42 & 88.988 & 6.684 \\
			CViT w/ Re-att & 90.45& 96.78 & 52.74& 66.42 & 62.13 & 65.87 & 58.17 & 65.72 & 88.989 & 6.684 \\
			CViT w/ MAS & 90.45 & 97.03 & 55.86 & 65.35 & 63.27 & 67.85 & 60.32 & 67.62 & 88.988 & 6.684 \\
			\midrule
			EViT  & 88.26 & 96.14 & 49.00 & 62.47 & 64.78 & 70.12 & 66.67 & 70.60 & 104.167 & 0.169 \\
			EViT w/ Re-att  & 88.12 & 96.38 & 52.54 & 63.27 & 65.31 & 70.46& 66.35 & 70.82 & 104.167 & 0.169 \\
			EViT w/ MAS & 88.44 & 96.66 & 57.30 & 64.80 & 66.26 & 70.71 & 66.39 & 71.34 & 104.167 & 0.169 \\
			\midrule
			CEViT & 93.67 & 98.36 & 44.24 & 65.29 & 66.14 & 75.55 & 62.61 & 67.51 & 90.640 & 0.174 \\
			CEViT w/ Re-att  & 93.85 & 98.82 & 47.14& 66.04 & 66.26 & 72.51 & 63.06 & 68.35 & 90.644 & 0.174 \\
			CEViT w/ MAS & 94.69 & 99.08 & 49.97 & 67.37 & 66.64 & 74.13 & 64.18 & 69.30 & 90.640 & 0.174 \\
			\bottomrule
			Ours & 94.76 & 98.93 & 65.09 &  73.21& 70.45&78.23 & 65.45 & 73.73 & 29.912& 9.799 \\
			Ours w/ Re-att & 95.32 & 99.03 & 66.31&  74.03& 71.27&79.52 & 68.21& 74.36 & 29.915& 9.802 \\
			Ours  w/ MAS & \textbf{97.13} & \textbf{99.70} & \textbf{66.68} & \textbf{75.32} & \textbf{72.13} & \textbf{80.01} & \textbf{71.24} & \textbf{78.77} &29.912  & 9.799\\
			\bottomrule
		\end{tabular*}%
		\label{tab:addlabel}%
	}
\end{table*}%
\begin{table*}[t!]
	\captionsetup{labelsep=none}
	\caption{ \ Ablation of the depth of transformer blocks. The best results are in bold.}
	\small
		\setlength{\tabcolsep}{0.5mm}{%
	\begin{tabular*}{\textwidth}{@{\extracolsep\fill}ccccccccccc@{}}
		\toprule
		\multirow{2}*{Loss function} & \multicolumn{2}{c}{FF++} & \multicolumn{2}{c}{Celeb-DF} & \multicolumn{2}{c}{DFDC} & \multicolumn{2}{c}{DF-1.0} & \multirow{2}*{\shortstack{Params\\(M)}}& \multirow{2}*{\shortstack{FLOPs\\(G)} }\\
		\cmidrule{2-4} \cmidrule{4-5} \cmidrule{5-7} \cmidrule{7-9}
		& ACC & AUC & ACC & AUC & ACC & AUC & ACC & AUC \\
		\midrule
		$L$=4 & 95.03 & 97.93 & 64.31 & 73.89 & 70.38 & 78.48 & 69.98 & 76.43 & 23.590 & 9.393 \\
		$L$=5 & 96.35 & 98.04 & 65.45 & 74.46 & 71.54 & 79.34 & 70.03 & 77.05 & 26.751 & 9.596 \\
		$L$=6 & \textbf{97.13} & \textbf{99.70} & \textbf{66.68} & \textbf{75.32} & \textbf{72.13} & \textbf{80.01} & \textbf{71.24} & \textbf{78.77} & 29.912& 9.799 \\
		$L$=7 & 96.85 & 98.32 & 65.54 & 74.27 & 72.06 & 79.43 & 70.32 & 77.67 & 33.073 &  10.002\\
		\bottomrule
	\end{tabular*}}
	\label{tab:61}%
\end{table*}

{\bfseries\setlength\parindent{0em} Effect of the number of transformer blocks.} We  conducted four experiments to study the effect of depths of transformer blocks. As Table~\ref{tab:61} shows, the performance of DTN gradually increases when the transformer blocks goes deeper. In detail, a nearly 1\% growth of AUC could be achieved by adding a transformer block, with an increase of 3.161M parameters and 0.203G FLOPs, which demonstrates that performance improvements could be realized with high computational efficiency as the depth of the transformer block increases. The AUC attains maximum as six transformer blocks are involved. We argued that MAS increases the diversity of attention maps with the depth growing, guiding the model to dig more general and comprehensive forgery traces.
\begin{figure}[t!]%
\includegraphics[width=\linewidth]{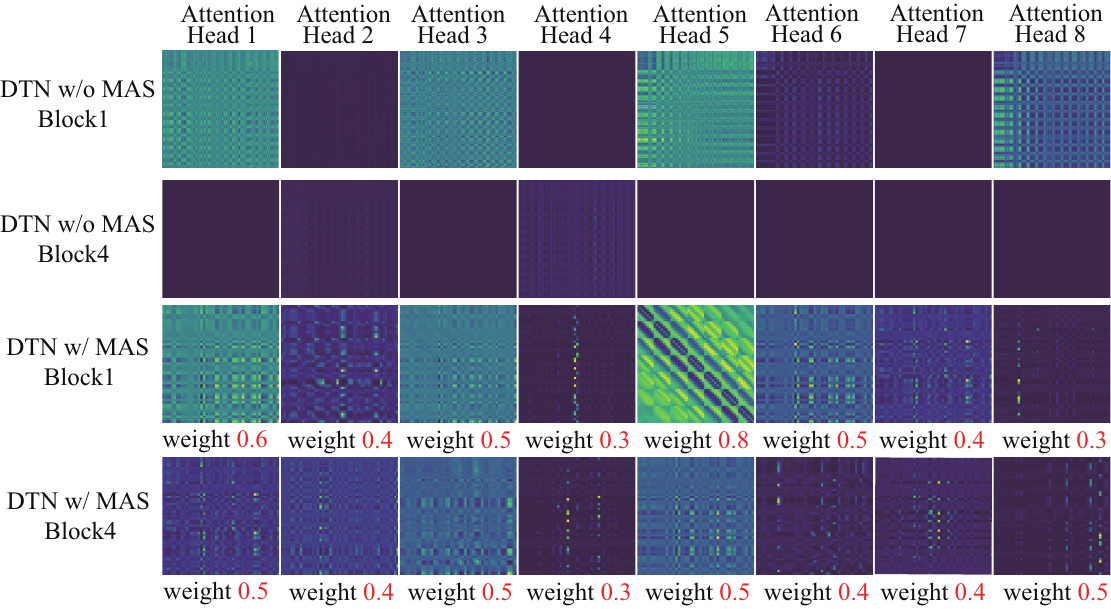}
\captionsetup{labelsep=none}
\caption{\ The visualization of attention maps across heads in various transformer blocks (w/o or w/ MAS).}\label{fig31}
\end{figure}

\begin{figure}[t!]%
\includegraphics[width=\linewidth]{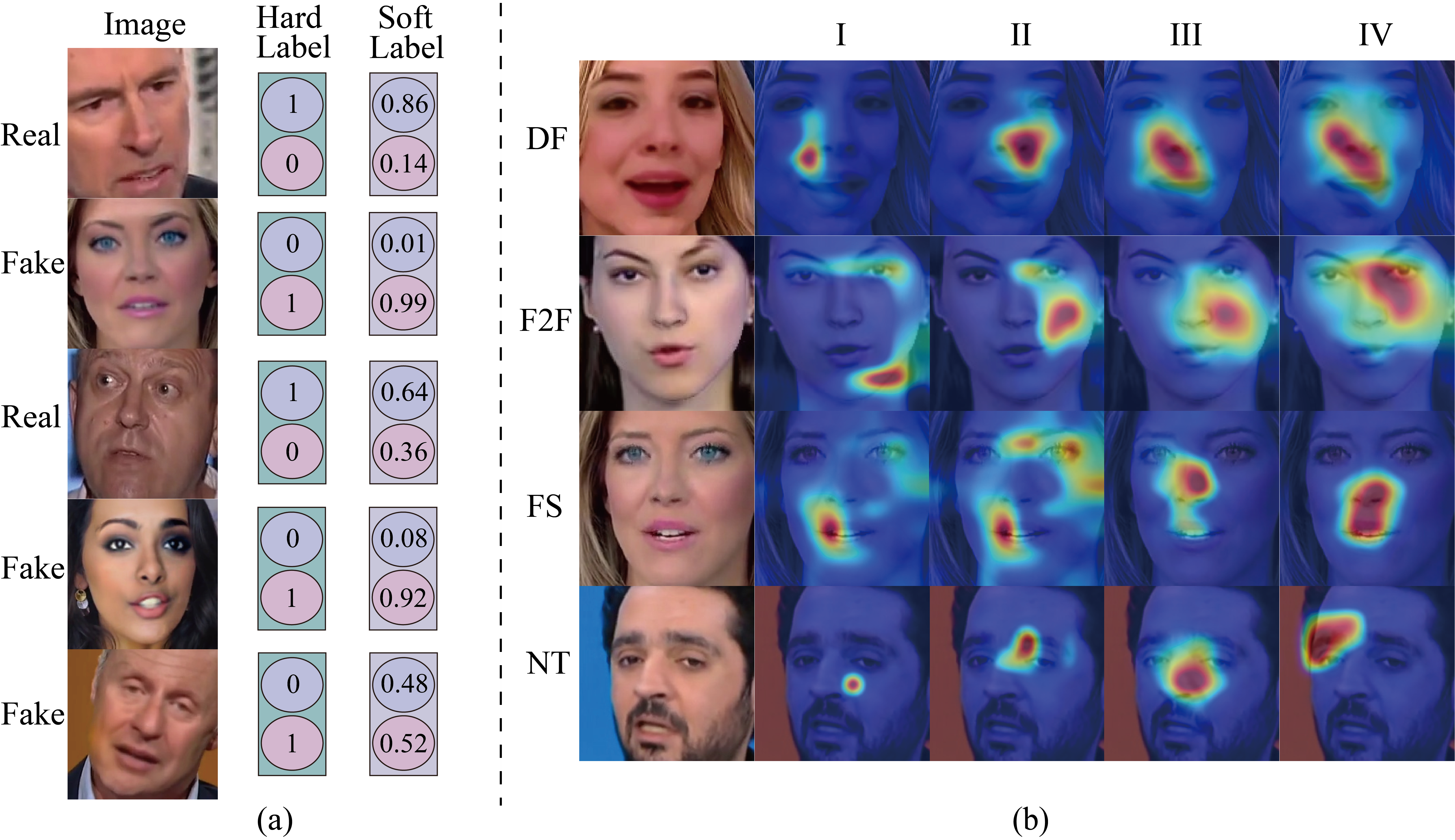}
	\captionsetup{labelsep=none}
\caption{ (a) The one-hot label visualizations of various examples from FF++. (b)The heatmap visualizations of different training settings on examples from FF++.}\label{fig7}
\end{figure}

{\bfseries\setlength\parindent{0em} Influence of MAS.} We further performed within-dataset and cross-dataset evaluations to demonstrate that our plug-and-play MAS module can generalize to different transformer-based models for performance gains, which are trained on FF++ and tested on FF++, Celeb-DF, DFDC, and DF-1.0. As Table~\ref{tab3} shows, it is evident that MAS boosts the generalizability of transformer-based models to unseen forgeries.  For example, due to the addition of MAS, the performance of ViT and CrossEfficientViT are improved by 1.1\% AUC and 1.8\% AUC, respectively, on the challenging DF-1.0 dataset. To further demonstrate the efficency of MAS, as illustrated in Figure~\ref{fig31}, we visualized attention maps across heads in different transformer blocks, when MAS is involved or not. The detailed value number of learnable weights in MAS is shown in Figure~\ref{fig31}. We noticed that attention maps across heads in the transformer block become more abundant, due to the introduction of MAS. Furthermore, the diversity of attention maps is improved as transformer blocks go deeper. It is assumed that MHSA in the transformer-based model hardly produces various attention maps, without additional losses. MAS realizes this by adaptively choosing salient attention heads and ignoring unimportant ones, supervising the transformer-based models to capture more diverse and general forgery representations. Compared to the Re-attention method that improves the diversity of attention maps via convolutional layers, MAS slightly introduces extra weights and computational costs when applied to other models.

\begin{figure}[ht!]%
	\includegraphics[width=\linewidth]{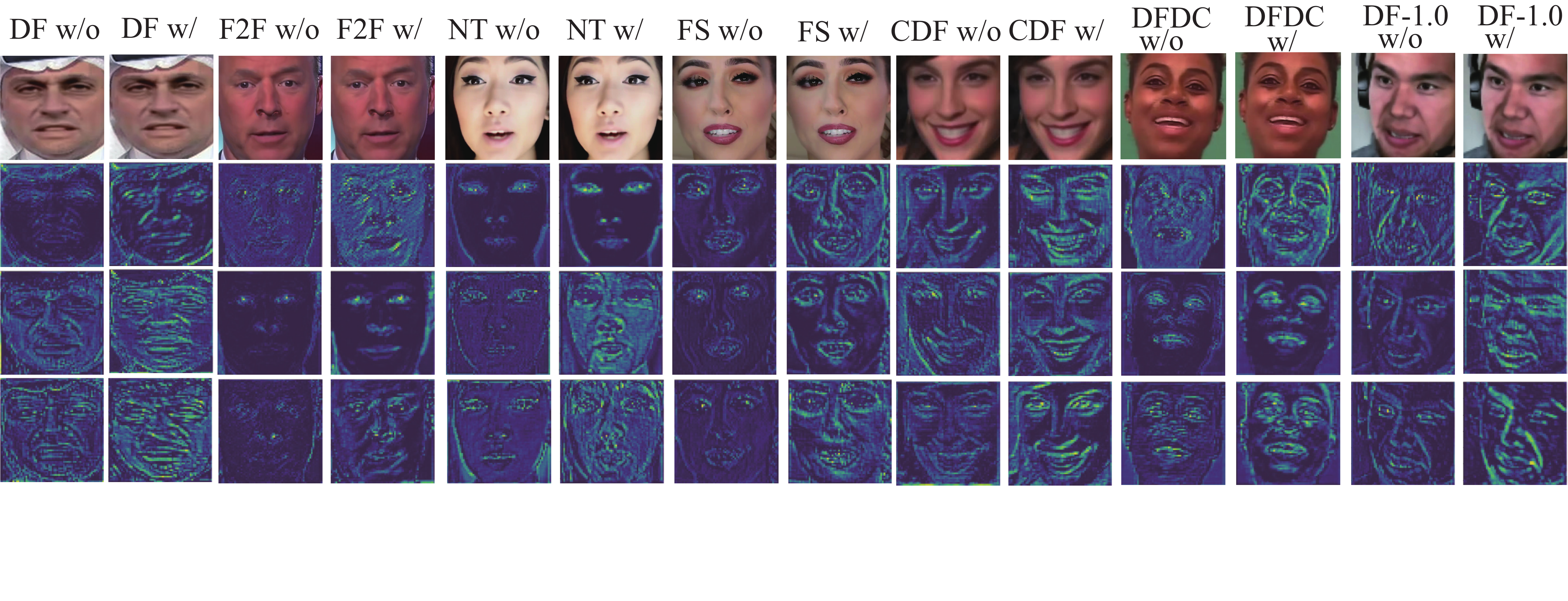}
	\centering
	\captionsetup{labelsep=none}
		\vspace{-3em}
	\caption{ \ The visualization of feature maps (w/o or w/ MAS). We tested models on FF++, Celeb-DF, DFDC, and DF-1.0 after training on FF++.}\label{fig51}

\end{figure}

As Figure~\ref{fig51} shows, we also visualized the ablated feature maps. The first row represents the RGB images. Every two columns display the same fake sample of various databases. We can see that MAS can push the network to learn long-range forgery traces and our MAS method performs well even on unseen manipulations.

\begin{table*}[t!]
	\centering
	\captionsetup{labelsep=none}
	\caption{\ The impacts of deepfake self-distillation learning on the CNN-based and transformer-based models. We use the network from the row above as a teacher to guide each student for learning. Student $i$ represents the network after the $i$-th iteration of learning. EViT denotes the EfficientViT model\label{tab6}}
	\small
	\setlength{\tabcolsep}{0.5mm}{%
		\begin{tabular*}{\textwidth}{@{\extracolsep\fill}lccccccccccc}
			\toprule
			\multirow{2}*{Method} & \multicolumn{2}{c}{ FF++} & \multicolumn{2}{c}{Celeb-DF} & \multicolumn{2}{c}{ DFDC} & \multicolumn{2}{c}{ DF-1.0} & \multicolumn{2}{c}{DFD} \\			
			\cmidrule{2-4} \cmidrule{4-5} \cmidrule{5-7} \cmidrule{7-9}\cmidrule{9-11}
			& ACC   & AUC   & ACC   & AUC   & ACC   & AUC   & ACC   & AUC   & ACC   & AUC \\
			\midrule
			Xception-Teacher & 94.08 & 96.51 & 54.24 & 65.86 & 58.77 & 66.95 & 54.76 & 67.03 & 76.84 & 85.20 \\
			Xception-Student1 & \textbf{95.80} & \textbf{97.56} & 57.57 & 66.53 & 60.08 & \textbf{67.44} & 58.93 & 71.67 & \textbf{78.43} & \textbf{86.65} \\
			Xception-Student2 & 95.74 & 97.52 & 58.99 & 66.47 & 59.97 & 67.21 & 59.68 & 73.29 & 77.55 & 86.03 \\
			Xception-Student3 & 95.77 & 97.43 & \textbf{60.81} & \textbf{67.36} & \textbf{60.10} & 66.89 & \textbf{61.60} & \textbf{74.16} & 77.92 & 85.92 \\
			Xception-Student4 & 95.61 & 97.45 & 59.50& 66.15 & 59.98 & 66.63 & 60.51 & 73.82 & 77.01 & 85.84 \\
			\midrule
			
			EViT-Teacher & 88.26 & 96.14 & 49.00 & 62.47 & 64.78 & 70.12 & 66.67 & 70.60 & 80.75 & 90.50 \\
			EViT-Student1 & 89.50 & 96.85 & 47.02 & 63.36 & 64.75 & 70.91 & 67.83 & 74.05 & \textbf{81.04} & 90.76 \\
			EViT-Student2 & 89.77 & 97.19 & 49.88 & \textbf{63.86} & 64.68 & 70.55 & 69.24 & 75.51 & 80.87 & \textbf{90.87} \\
			EViT-Student3 & 89.83 & 97.31 & \textbf{50.16} & 62.26 & \textbf{65.93} & \textbf{71.00 } & \textbf{70.95} & \textbf{77.95} & 79.19 & 88.86  \\
			EViT-Student4 & \textbf{90.07} &\textbf{ 97.52} & 49.08 & 62.56& 63.68 & 70.85 & 70.03 & 77.62 & 80.88 & 90.31  \\
			\midrule
			
			DTN-Teacher & 95.20 & 99.17 & 64.23 &73.72 & 70.39 & 78.01 & 68.03 & 75.36 & 85.06 & 94.83\\
			DTN-Student 1 & 95.91 & 99.55 & 66.03 &74.81 & 70.71 & 78.72 & 67.02 & 76.68 & 85.48 & 96.34 \\
			DTN-Student 2 & 96.14 & 99.69 & 66.17 & 75.12 & 70.93 & 79.90 &\textbf{ 71.35} & 78.54 & 86.52 & 97.53 \\
			DTN-Student 3 & 97.13 & 99.70 & \textbf{66.68} & \textbf{75.32} & \textbf{72.13} & \textbf{80.01} & 71.24 & \textbf{78.77} & \textbf{87.32} & \textbf{97.60} \\
			DTN-Student 4 & \textbf{97.35} & \textbf{99.81} & 65.32 & 74.05 & 72.01& 79.83 & 70.36 & 78.42 & 87.23 & 97.54 \\
			\bottomrule
		\end{tabular*}%
	}
\end{table*}%

{\bfseries\setlength\parindent{0em} Effect of deepfake self-distillation.} We further investigated the importance of deepfake self-distillation learning on CNN-based and transformer-based networks, by training on FF++ and testing on FF++, DFDC, Celeb-DF, DF-1.0, and DFD. In Table~\ref{tab6}, we noticed that AUC increased by 1.1\%, 0.7\%, 1.3\%, and 1.5\% on Celeb-DF, DFDC, DF-1.0, and DFD, respectively, from the teacher to student 1 model, which verifies that it is helpful to introduce inter-class similarity knowledge. We can see that when the student becomes the teacher of the subsequent generation, the performance of next-generation students can be consistently improved, i.e., the AUC of student 2 (78.5\%) on DF-1.0 is 1.8\% higher than that of student 1 (76.7\%). However, the model generalization ability tends to decline after student 3, so we chose student 3 as our model. Concerning CNN-based models, it is noticed that the deepfake self-distillation learning strategy is also beneficial. 
\begin{figure}[t!]%
	\includegraphics[width=\linewidth]{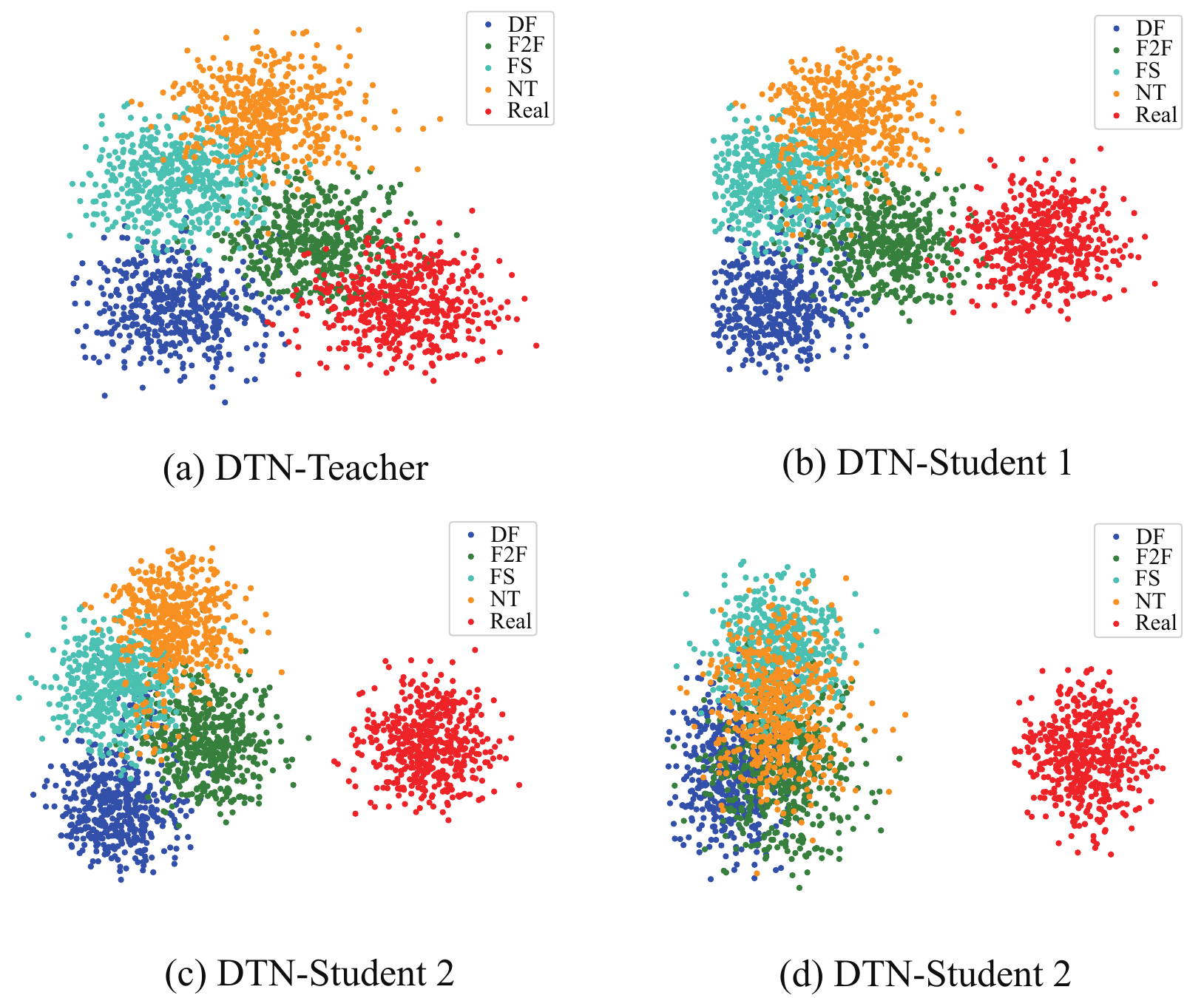}
	\centering
	\captionsetup{labelsep=none}
	\caption{ \ The t-SNE visualizations of various models. We performed a random selection of 500 images for each category from the testing dataset of FF++, consisting of both natural faces and four types of manipulated ones.}\label{tsne}
\end{figure}

\subsection{Visualization}
{\bfseries\setlength\parindent{0em} Visualization of sample labels.}
To investigate the correctness of the one-hot label, we randomly selected some samples from FF++, and visualized the one-hot labels, including hard label and soft label yielded by the DTN teacher 3 model. As Figure~\ref{fig7} (a) shows, the DTN teacher 3 model generates correct soft labels in most cases, such as the discriminative ones [0.01, 0.99] and [0.08, 0.92], which can promote the student model to learn richer and more authentic information.

{\bfseries\setlength\parindent{0em} Visualization of feature distribution.} To further investigate the impact of deepfake self-distillation on feature distribution, we utilize t-SNE to visualize the features extracted from samples in FF++. As is displayed in Figure~\ref{tsne}, features of various manipulations encoded by the teacher are clustered in different domains. With the increase of iterations, the samples of different forgery faces start to cluster together, and finally become a single cluster. The boundary between real and fake faces, based on the distribution of our features, is much more evident. It proves that our network can extract more general and common embeddings for different forgery faces.

{\bfseries\setlength\parindent{0em} Visualization of heatmap.}
To further study the effectiveness of the proposed model, we used the Grad-CAM \citep{Selvaraju_2017_ICCV} to visualize the heatmap produced by different settings. In Figure~\ref{fig7} (b), we showed the class activation mapping (CAM) of four samples for four models. Each row displays a fake face generated by various manipulation approaches, i.e. DF, F2F, FS, and NT. The second to fifth columns display heatmaps for four models listed in Table~\ref{tab2}: (I) the baseline model; (II) LEVT + baseline; (III) baseline + MoE + LEVT; and (IV) the full architecture, consisting of MoE, LEVT, and MAS modules. LEVT (II) can capture more long-range manipulation traces than baseline (I), which suggests that our LEVT module explores locally-enhanced global relations. In comparison to (II), MoE (III) further mines more potential forgery regions. MAS (IV) boosts these manipulation areas by learning more abundant representations when compared to (III). 	

\section{Conclusion}\label{sec5}
In this paper, we propose an innovative method for FFD, which aims to provide detectors with soft labels for supervision, and address the limitation of attention collapse to capture general and diverse forgery traces. On one hand, we design the deepfake self-distillation learning scheme, which generates the soft labels automatically via the STG module, and can be applied to any detectors to enhance the generalization performance. On the other hand, a plug-and-play MAS module is designed to avoid attention collapse via adaptively selecting attention maps, which can be integrated into any transformer-based model with only a slight growth in computational costs. We design the MoE module to mine the robust forgery traces, and the LEVT module to learn locally-enhanced global features. 

{\bfseries\setlength\parindent{0em} Limitations.}  Although we provide soft tags automatically using the pretrained teacher model, we need to carefully select the samples with incorrect predictions and those with similar true and false probabilities. For the former limitation, we can increase the weights of incorrectly predicted samples so that the model can iteratively learn more about them. For the latter limitation,  we can perform data agumentation (such as rotation, flipping, scaling, etc.) on ambiguous samples to generate more training instances, to help the model learn the general and discriminative features. Furthermore, in the future, we propose to cope with the challenge of detecting high-quality deepfake images generated by diffusion models, using advanced technology such as the novel state space model architecture Mamba.

{\bfseries\setlength\parindent{0em}Acknowledgements} This work was supported by the National Natural Science Foundation of China under Grant 82261138629; Guangdong Basic and Applied Basic Research Foundation under Grant 2023A1515010688 and Shenzhen Municipal Science and Technology Innovation Council under Grant JCYJ202205311014
12030.



  \bibliographystyle{elsarticle-num-names} 
  \bibliography{dtn_bib}





\end{document}